\documentclass[sigplan,twocolumn]{acmart}
\renewcommand\footnotetextcopyrightpermission[1]{}
\usepackage{makecell}
\usepackage{subcaption}
\usepackage{caption}   
\usepackage{multirow}
\usepackage{pifont}
\begin{document}

\newcommand{\modelname}{HOBBIT }
\newcommand{\modelnamenospace}{HOBBIT}
\newcommand{\modelnameshort}{HB }
\newcommand{\modelnameshortnospace}{HB}

\title{\includegraphics[width=0.05\textwidth]{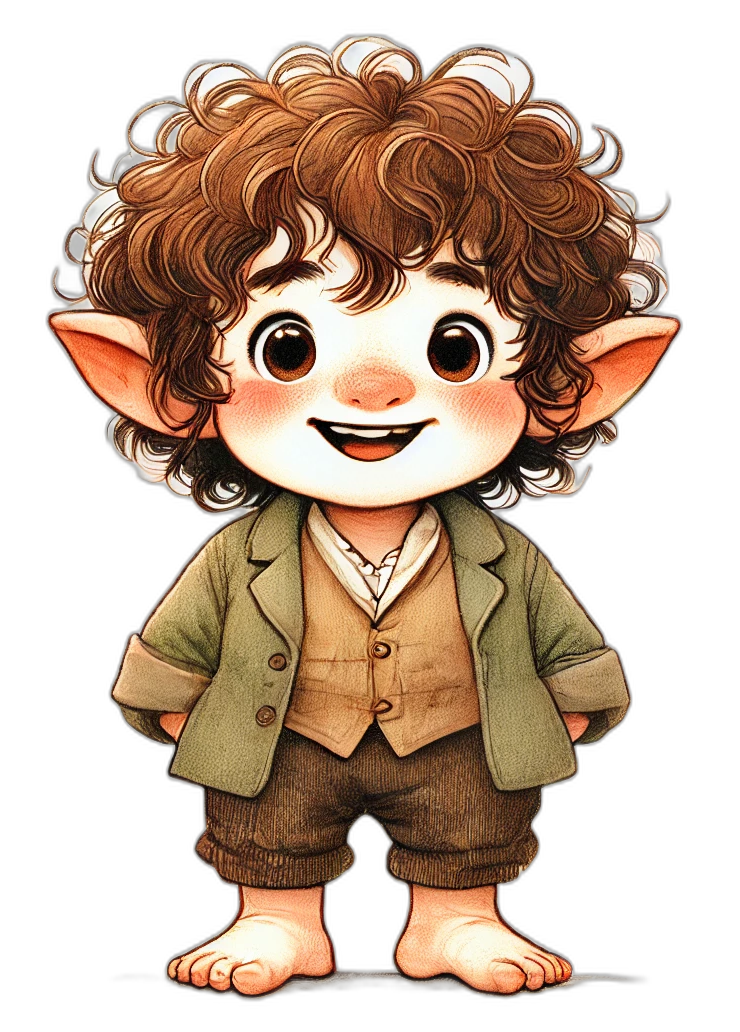} HOBBIT: A Mixed Precision Expert Offloading System for Fast MoE Inference}

\author{
    Peng Tang\textsuperscript{1,*}, 
    Jiacheng Liu\textsuperscript{2,*}, 
    Xiaofeng Hou\textsuperscript{1,$\dagger$}, 
    Yifei Pu\textsuperscript{1}, 
    Jing Wang\textsuperscript{1}, 
    Pheng-Ann Heng\textsuperscript{2},
    Chao Li\textsuperscript{1,$\dagger$}, 
    Minyi Guo\textsuperscript{1}\\
    \textsuperscript{1}\textit{Department of Computer Science and Engineering, Shanghai Jiao Tong University, Shanghai, China}\\
    \textsuperscript{2}\textit{Department of Computer Science and Engineering, The Chinese University of Hong Kong, China}\\
    \textsuperscript{*} Equal contribution, \textsuperscript{$\dagger$} Corresponding authors\\
}

\renewcommand{\shortauthors}{Peng Tang et al.}
\renewcommand{\shorttitle}{HOBBIT: A Mixed Precision Expert Offloading
System for Fast MoE Inference}

\begin{abstract}
The Mixture-of-Experts (MoE) architecture has demonstrated significant advantages in the era of Large Language Models (LLMs), offering enhanced capabilities with reduced inference costs. 
However, deploying MoE-based LLMs on memory-constrained edge devices remains challenging due to their substantial memory requirements. 
While existing expert-offloading methods alleviate the memory requirements, they often incur significant expert-loading costs or compromise model accuracy.
We present \modelnamenospace, a mixed precision expert offloading system to enable flexible and efficient MoE inference. Our key insight is that dynamically replacing less critical cache-miss experts with low-precision versions can substantially reduce expert-loading latency while preserving model accuracy. \modelname introduces three innovative techniques that map the natural hierarchy of MoE computation: (1) a token-level dynamic expert loading mechanism, (2) a layer-level adaptive expert prefetching technique, and (3) a sequence-level multidimensional expert caching policy. These innovations fully leverage the benefits of mixed-precision expert inference.
By implementing \modelname on top of the renowned LLM inference framework Llama.cpp, we evaluate its performance across different edge devices with representative MoE models. The results demonstrate that \modelname achieves up to a 9.93x speedup in decoding compared to state-of-the-art MoE offloading systems.
\end{abstract}

\maketitle
\section{Introduction}

The rapid explosion of Large Language Models (LLMs) has led to their widespread application across various fields~\cite{zhao2023survey}. Beyond deploying LLM in cloud-based data centers,  there is a growing demand to deploy these models at the edge to address issues like high latency, privacy concerns, and dependence on stable network connections inherent in centralized approaches~\cite{friha2024llm}. Consequently, there is an increasing need to run LLMs on edge devices, bringing intelligence closer to the end-user. Nowadays, both academia~\cite{chu2024mobilevlm, zhang2024tinyllamaopensourcesmalllanguage, xue2024powerinfer} and industry~\cite{qualcomm, huaweai, appleai} are actively accelerating the deployment of LLMs at the edge.

In recent years, the Mixture of Experts (MoE) architecture~\cite{shazeer2017outrageously} has emerged as a promising approach to enhance LLM capabilities by enabling significant model size expansion while maintaining computational efficiency~\cite{jiang2024mixtral, abdin2024phi, DeepSeekV2, qwen_moe, Switchtransformers}.  However, MoE-based LLMs demand substantial GPU memory for parameter storage. For instance, the Mixtral-8x7B model~\cite{jiang2024mixtral}, despite activating only 14 billion parameters per token, requires 87GB of memory to store its complete set of 45 billion parameters. This poses significant deployment challenges on memory-constrained edge devices, such as the NVIDIA Jetson AGX Orin with its 32GB memory capacity. To address this limitation, expert-offloading techniques have been developed to enable the execution of these large-scale models on memory-limited devices by exploiting the sparse activation patterns inherent in MoE architectures.

In essence, expert-offloading techniques primarily store all non-expert weights and a subset of important experts in GPU memory (referred to as the "expert cache"), while offloading other experts to CPU memory or SSD (referred to as "next-level memory"). When the required experts are not available in the expert cache, they are loaded from next-level memory into the cache, evicting some existing experts. However, due to limited memory bandwidth, loading an expert from next-level memory introduces significant latency, which can severely slow down inference. While existing systems optimize expert-offloading with various methods, they still face several limitations, as outlined below.

\textbf{Inflexible and aggressive optimizations of expert loading.}  
When an expert cache miss occurs, directly loading a missing expert incurs significant latency.  To mitigate this, EdgeMoE~\cite{yi2023edgemoe} employs different quantization levels for various experts to reduce I/O costs and  AdapMoE~\cite{adamoe} skips certain experts to decrease loading costs. 
However, these approaches have notable limitations. EdgeMoE's static approach determines optimal bit widths based on specific dataset profiling, leading to inflexibility across diverse environments and potential accuracy impacts. This method becomes particularly complex when dealing with different models, especially as the number of experts increases. Conversely, AdapMoE's aggressive expert-skipping strategy can cause substantial accuracy degradation, particularly with small top-$k$ values (e.g., $k=2$ in Mixtral-8x7B).

\textbf{Limited benefits of expert prefetching.} 
To reduce the waiting time for required experts, prefetching is a valuable technique that overlaps expert loading with GPU computation. However, since MoE models only need the top-$k$ experts for the next layer, accurately predicting these top-$k$ experts is crucial. MoE-Infinity~\cite{xue2024moe} addresses this by prioritizing expert activation ratios for prefetching. MoE-Offloading~\cite{eliseev2023fast} uses the gate inputs from the current layer as inputs for the next layer to predict the required experts. Pre-gated MoE~\cite{hwang2024pre} modifies the model structure by introducing a pre-gate function to determine the next layer's required experts in the current layer. Although prefetching can overlap expert-loading with GPU computation, these prediction methods offer limited benefits because the expert-loading cost is typically much greater than the GPU computation cost in the inference process of MoE-based LLMs.

\textbf{Inefficient management of expert cache.}
Given the sparse activation and temporal locality characteristics of experts, designing an appropriate cache replacement policy to manage the expert cache can significantly improve the expert cache hit ratio, reducing the need to load experts from next-level memory and thereby speeding up inference. For instance, EdgeMoE~\cite{yi2023edgemoe} and MoE-Infinity~\cite{xue2024moe} utilize the least frequently used (LFU) policy, while MoE-Offloading~\cite{eliseev2023fast} adopts the least recently used (LRU) policy. Although these approaches outperform random replacement policies, they are not fully optimal, as they fail to account for the unique characteristics of different models, which requires more tailored strategies to manage the expert cache efficiently.

To address the above challenges, we propose \modelnamenospace, a system designed to accelerate expert loading across three levels of MoE computation. It significantly accelerates MoE-based LLM inference on memory-limited devices compared to existing systems by utilizing mixed precision expert inference. Our key contributions are as follows:

\begin{itemize}

    \item 
    We propose a token-level dynamic expert loading mechanism that reduces latency through low-precision replacement of less critical cache-miss experts, maintaining accuracy and flexibility.

    \item We develop a layer-level adaptive expert prefetching technique with high prediction accuracy and minimal penalties, leveraging mixed-precision prefetching to optimize computation-communication overlap.

    \item We introduce a sequence-level multidimensional expert caching policy that combines model-specific strategies with mixed-precision features to efficiently manage the expert cache and minimize miss penalties across different models.

    \item We implement \modelname on top of Llama.cpp with 8,000 additional lines of C++/C code, and evaluate it on two popular MoE-based LLMs across two memory-limited platforms, demonstrating up to 9.93x speedup in decoding over state-of-the-art systems.
\end{itemize}

\section{Background and Motivation}
\subsection{Background}

\begin{figure}[t]
    \centering
    \subfloat[Dense FFN Layer]{\includegraphics[width=1\linewidth]{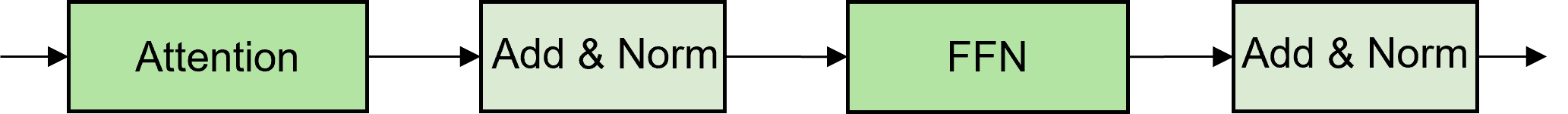}}
    \hfill
    \subfloat[Sparse MoE Layer]{\includegraphics[width=1\linewidth]{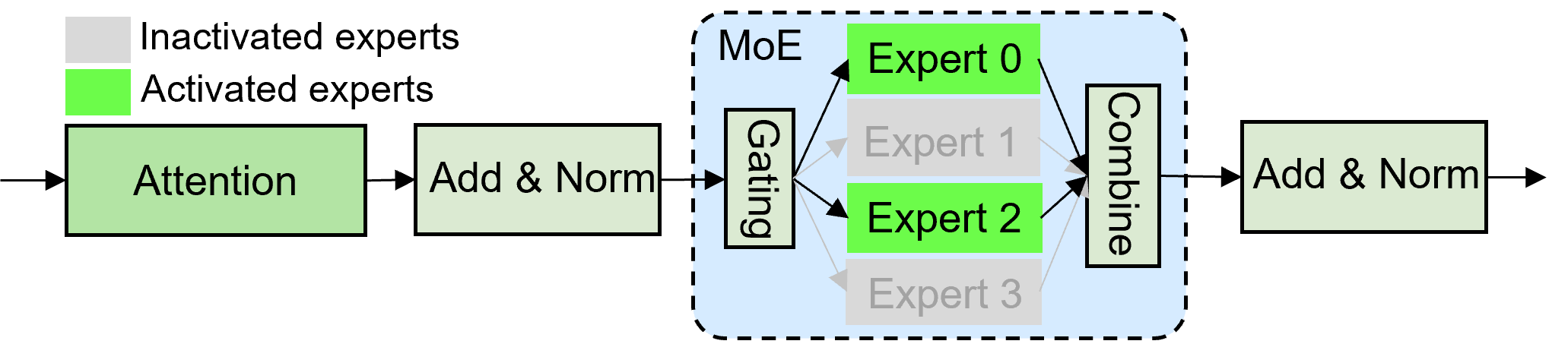}}
    \caption{Comparison between different LLM architectures.}
  \label{fig:moelayer}
\end{figure}

\textbf{Spare MoE Layers}. Due to the effectiveness of the MoE architecture~\cite{articleJacobs}, numerous MoE-based models~\cite{MetaPi,NIPS20009fdb62f9,Eigen2013LearningFR} have emerged. In this work, we focus on the most widely used sparse MoE layer~\cite{shazeer2017outrageously}, which employs FFNs as experts. As shown in Figure~\ref{fig:moelayer}, unlike dense layers, the MoE layer uses a gating function to select the $K$ most relevant experts (2 in the figure) for each input token, aggregating their outputs. This approach mimics specialized processing in different brain regions, enhancing model performance without increasing computational complexity.
For an input $x$, the output $y$ of the MoE module can be formulated as:
\begin{equation}\label{eq:moe-compute}
    y = \sum_{i=1}^K G(x)_{e_{i}}E_{e_{i}}(x)
\end{equation}
where $e_{i}$ is the $i$-th selected expert in the current layer, $G(x)_{e_{i}}$ represents the gating weight of expert $e_{i}$, and $E_{e_{i}}(x)$ is the output of expert $e_{i}$. 
The gating function $G(x)$ is typically implemented using a linear layer followed by a Top-k operation~\cite{jiang2024mixtral, DeepSeekV2, qwen_moe, Switchtransformers}.
By stacking MoE layers with multiple experts, LLMs can scale to massive sizes, improving performance while maintaining computational efficiency.

\noindent \textbf{Expert Offloading}. Parameter-offloading techniques typically transfer part of the model's parameters to CPU memory or SSDs when GPU memory is insufficient~\cite{deepspeedinference}. However, most offloading systems, such as Zero-Infinity~\cite{rajbhandari2021zero} and Accelerate~\cite{accelerate}, are designed for dense LLMs and load model parameters layer-by-layer on demand. This approach overlooks the sparse activation nature of MoE models, resulting in substantial latency. For instance, loading a layer of the Mixtral-8x7B model from CPU memory via a PCIe 4.0 link (32GB/s) takes approximately 80ms, while computing the same layer on an RTX 4090 GPU requires only about 3ms.

\begin{figure}[t]
    \centering
    \subfloat[Hardware Architecture]{\includegraphics[width=0.65\linewidth]{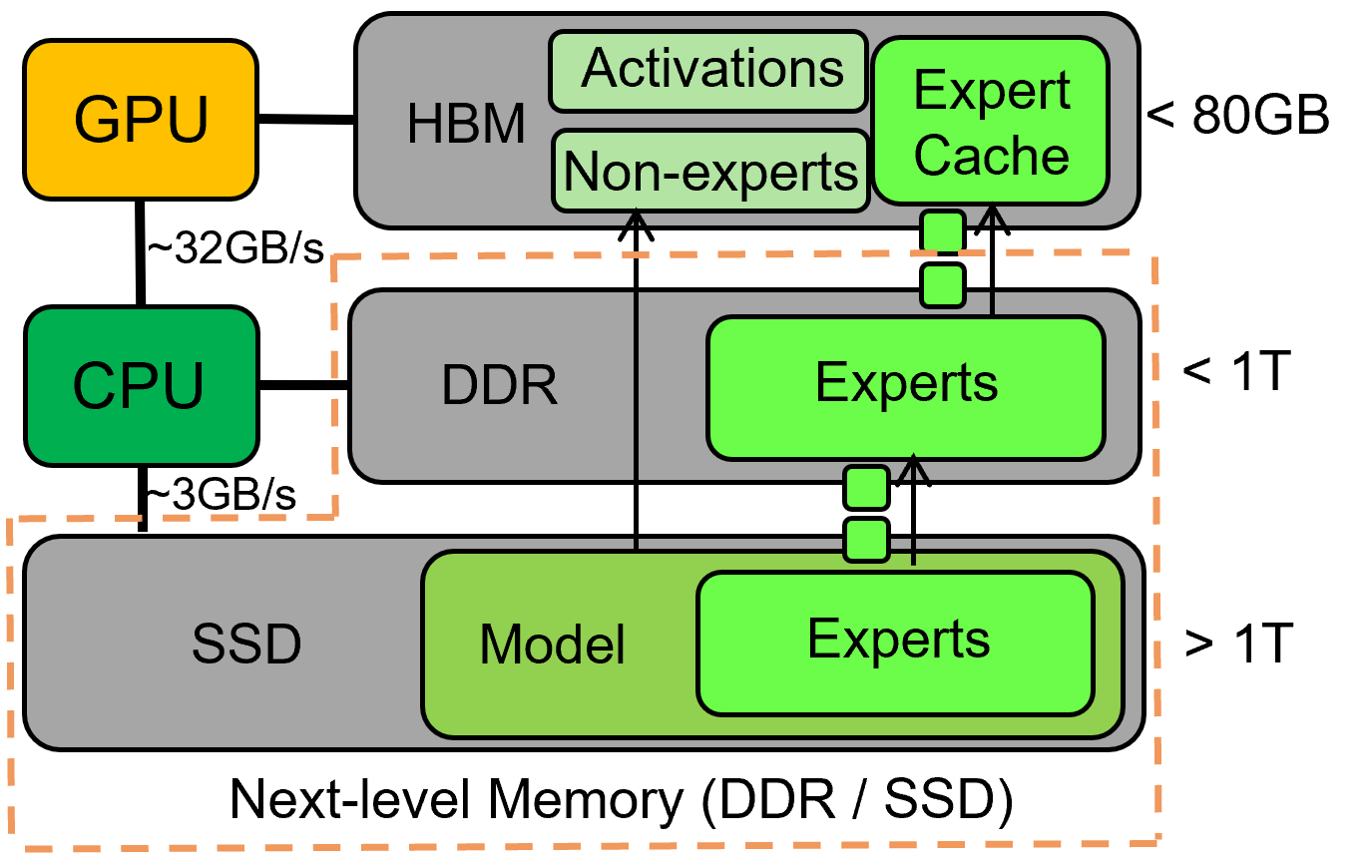}}
    \subfloat[MoE Parameters]{\includegraphics[width=0.34\linewidth]{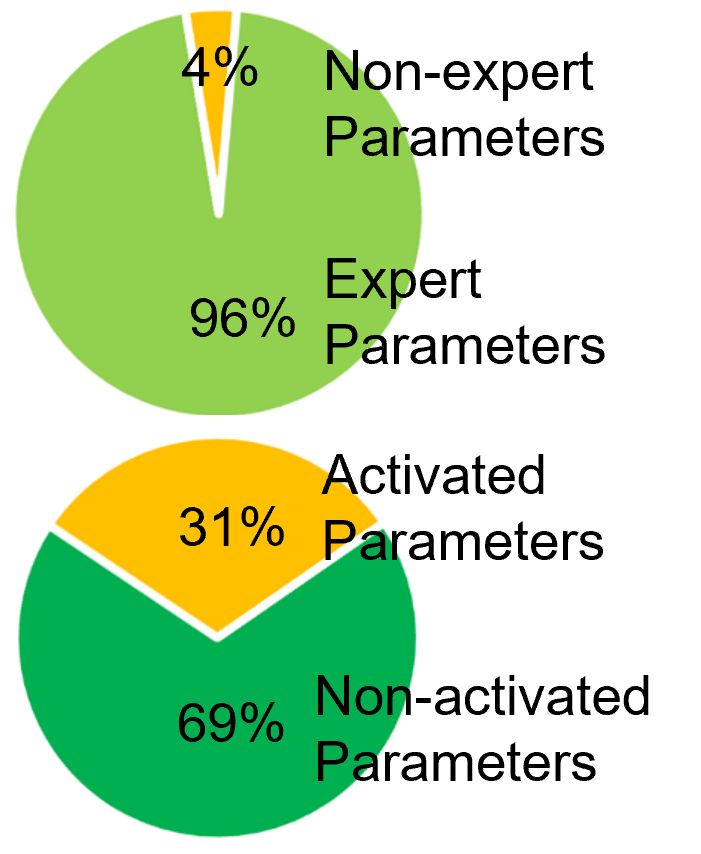}}
    \caption{Expert-offloading on hardware architecture and model parameter distribution for Mixtral-8x7B.}
  \label{fig:hardware}
\end{figure}

To address the latency issue of MoE models when using parameter offloading, some studies have developed expert-offloading, a specialized form of parameter-offloading tailored to the sparse activation characteristic of MoE~\cite{eliseev2023fast,kamahori2024fiddler,xue2024moe}. As shown in Figure~\ref{fig:hardware}-(a), this technique typically considers two levels of hardware memory: GPU memory stores all non-expert weights, a subset of "hot experts" (expert cache), and internal activations, while other experts are offloaded to CPU memory or SSD and loaded on demand. This approach is effective because each token requires all non-expert weights but only a fraction of experts. Figure~\ref{fig:hardware}-(b) illustrates this efficiency using the Mixtral-8x7B model as an example: non-expert weights constitute only 4\% of the model, and just 31\% of the parameters are activated per token. By leveraging this sparse activation pattern, expert-offloading significantly reduces GPU memory requirements while maintaining model functionality, making it possible to deploy large MoE models on memory-constrained devices.

Despite the effectiveness, existing expert-offloading techniques still incur high latency due to on-demand loading. 
While some of the works focus on optimizing prefetching techniques and cache replacement policies to accelerate inference speed, they remain constrained by the significant cost of expert loading during cache misses.

\subsection{Motivations}\label{sec:moti}

\begin{figure}[t]
    \centering
    \captionsetup[subfloat]{skip=-3pt}
    \subfloat[\centering Expert loading accounts for the majority of latency]{\includegraphics[width=0.50\linewidth]{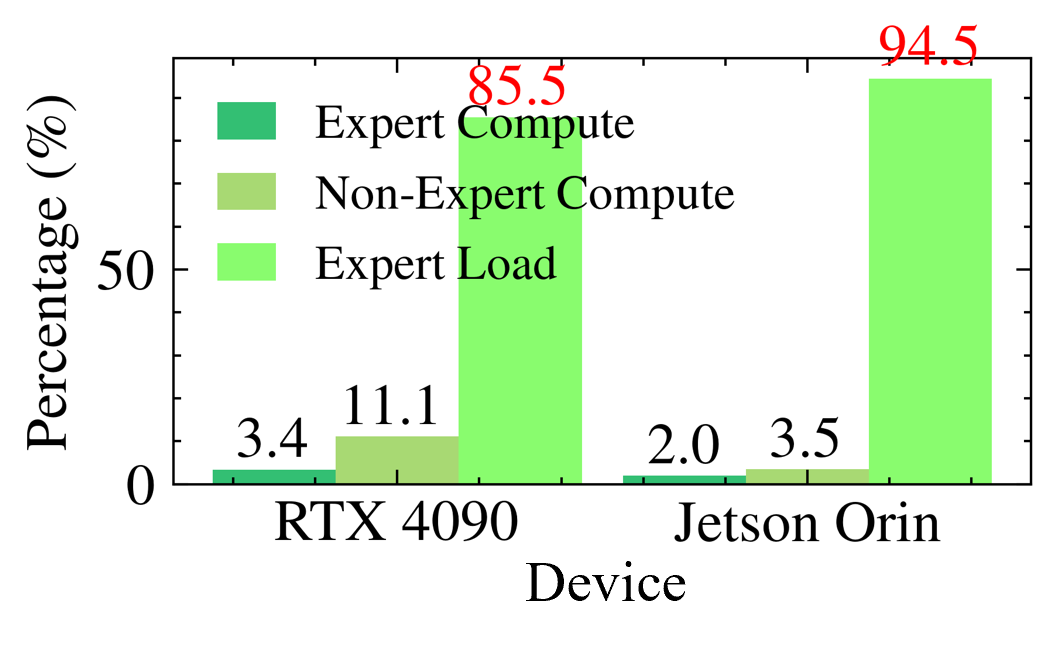}}
    \subfloat[\centering Low-precision expert replacement preserves accuracy]{\includegraphics[width=0.50\linewidth]{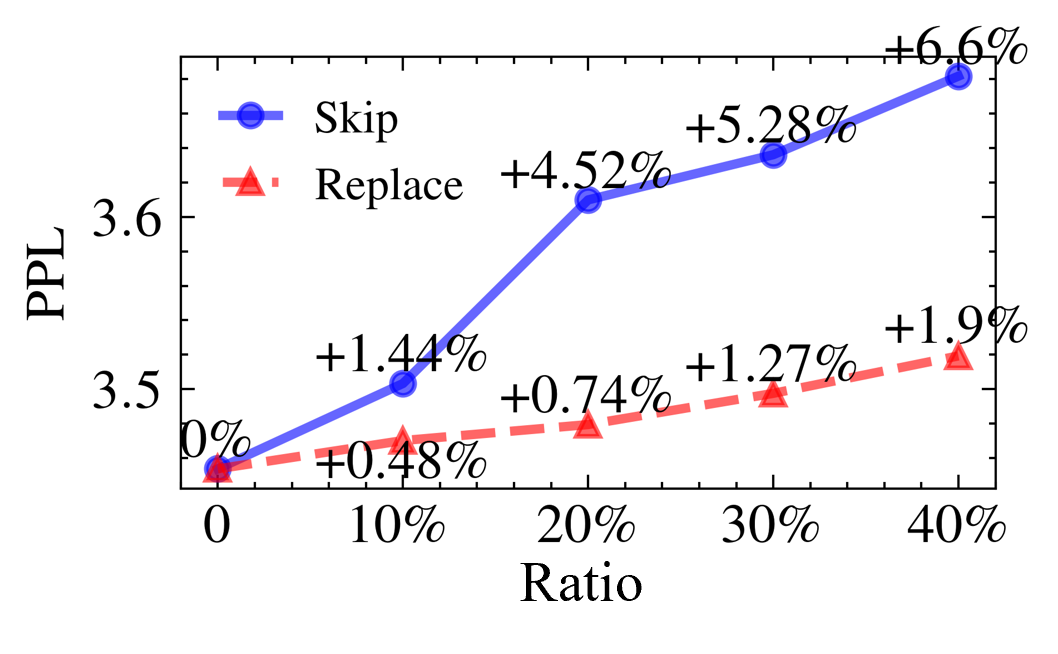}}
    \caption{Analysis of expert loading acceleration chances.}
  \label{fig:motivations}
\end{figure}

We identify two key observations that motivate our work:

\noindent \textbf{Expert loading dominates inference cost.}  
To quantify the bottlenecks in MoE model inference, we measured the time costs of different operations when running a Mixtral-8x7B layer on two memory-limited edge devices: an RTX 4090 (representing an edge server) and a Jetson Orin (representing an end device). As shown in Figure~\ref{fig:motivations}-(a), expert loading dominates the total inference time, consuming approximately 85.5\% on the RTX 4090 and 94.5\% on the Jetson Orin, while computation accounts for only a small fraction. While prefetching is commonly used to accelerate offloading by overlapping computation with data loading, its benefits are severely limited in MoE models due to this disproportionate time distribution. Some researchers have attempted to address this by employing dynamic gating to limit the number of experts loaded~\cite{adamoe,li2023adaptive}. However, this approach comes with significant accuracy trade-offs. As shown in Figure~\ref{fig:motivations}-(b), the "Expert Skip" method results in notable degradation of model performance, with a 10\% expert skip rate causing more than a 1\% increase in perplexity (PPL).

\noindent \textbf{Mixed precision expert preserves model accuracy.} Quantization is an effective method for reducing model parameter size, but directly quantizing the entire model can result in substantial accuracy loss. In MoE models, different experts have varying levels of importance~\cite{yi2023edgemoe, adamoe, kong-etal-2024-swapmoe}, so quantizing only the less important experts minimally impacts accuracy. As shown in Figure~\ref{fig:motivations}-(b), compared to skipping some experts, replacing them with low-precision versions better maintains model accuracy, and the gap between skipping and replacing grows as the ratio increases. In particular, when fewer than 20\% of the experts are quantized, model performance declines by no more than 1\%. Thus, applying quantization to low-importance experts in expert-offloading techniques can significantly reduce expert-loading cost. Specifically, if a required expert is not available in GPU memory and its importance is low, we can fetch a lower-precision version to replace it, thereby greatly reducing loading time. For instance, replacing a float16 expert with an int4 version can achieve up to a 4x speedup in the loading process. 

These observations motivate the need for a system that can dynamically manage expert precision during inference while maintaining model accuracy.

\begin{figure*}[t]
    \centering
    \includegraphics[width=\linewidth]{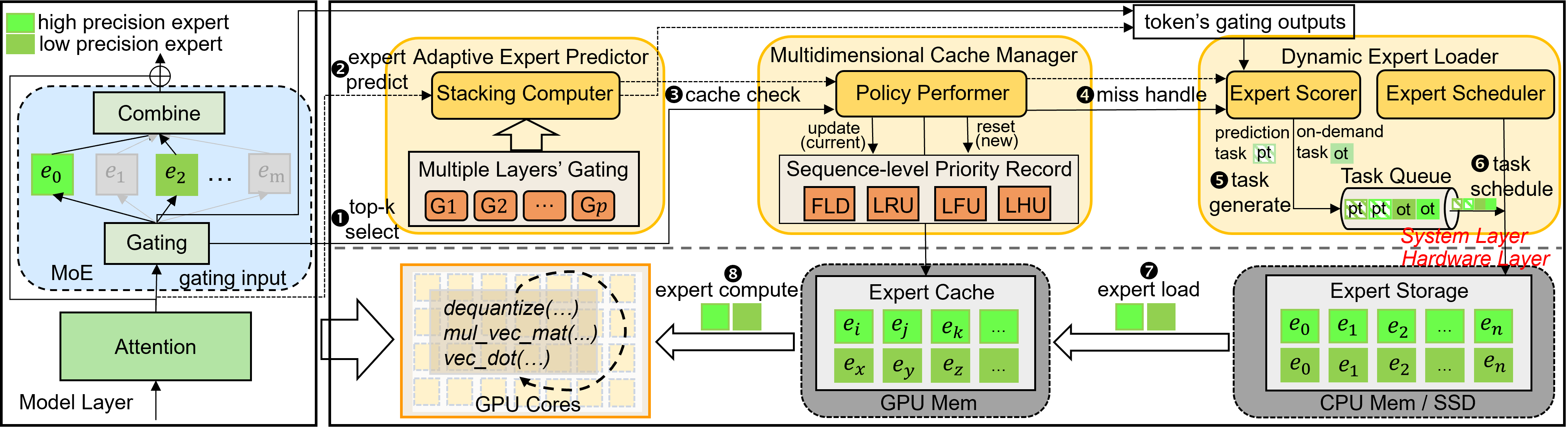}
    \caption{System overview of \modelnamenospace.}
    \label{fig:overview}
\end{figure*}

\section{\modelname System}

\subsection{Overview of \modelname}

\modelname is a mixed precision expert offloading system designed for the inference of MoE-based LLMs on memory-limited devices. 
It incorporates three-level innovations:
(i) a token-level dynamic expert loading mechanism that selects an appropriate precision expert from CPU memory or SSD through gating networks; 
(ii) a layer-level adaptive expert prefetching technique that provides highly accurate prefetching decisions for subsequent layers; and (iii) a sequence-level multidimensional expert caching policy that combines multiple cache replacement strategies along with the unique features of the mixed precision experts. The three-level design of \modelname directly maps to the natural hierarchy of MoE computation, ensuring comprehensive optimization while avoiding redundant granularities. %

As shown in Figure~\ref{fig:overview}, \modelname consists of three main modules built upon these mechanisms: Dynamic Expert Loader, Adaptive Expert Predictor, and Multidimensional Cache Manager. 
The Dynamic Expert Loader implements the dynamic loading mechanism to generate loading tasks for cache-miss experts and load corresponding precision experts. The Adaptive Expert Predictor leverages the adaptive prefetching technique to predict experts required for subsequent layers. The Multidimensional Cache Manager employs the proposed multidimensional caching policy to manage experts stored in GPU memory.

When executing a MoE layer on the GPU, the system first \ding{182} selects the top-k required experts (referred to as on-demand experts) for MoE computation based on the gating outputs. Simultaneously, the Adaptive Expert Predictor \ding{183} predicts the experts needed for subsequent layers (referred to as prediction experts) using its Stacking Computer, based on the current gating input. The Multidimensional Cache Manager then \ding{184} checks if the required experts are present in the expert cache and updates (for the current processing sequence) or resets (for a new coming sequence) the priority record with its Policy Performer. If all on-demand experts are present in the cache, \ding{189} the expert computation is performed on the GPU cores. 

If any on-demand or prediction experts are missing from the cache, the Dynamic Expert Loader uses the Expert Scorer to \ding{185} handle the cache miss based on the gating outputs of the current processing token. The Expert Scorer dynamically \ding{186} generates the corresponding loading tasks with varying precision requirements, adding them to the Task Queue. The Expert Scheduler module in the Dynamic Expert Loader \ding{187} then fetches tasks from the Task Queue and \ding{188} loads the corresponding experts from the Expert Storage into the Expert Cache. If necessary, the Multidimensional Cache Manager will replace older experts in the cache based on the proposed caching policy. The system waits for all on-demand expert loading tasks to complete before \ding{189} computing the outputs of the experts for the MoE module  and advancing  to the next layer. 
This process efficiently handles expert cache misses and accelerates inference by reducing expert-loading costs through the use of adaptive precision experts.

\subsection{Token-level Dynamic Expert Loading}

\begin{figure}[t]
    \centering
    \captionsetup[subfloat]{skip=-2pt}
    \subfloat[\centering Relationship between expert output and gating output]{\includegraphics[width=0.49\linewidth]{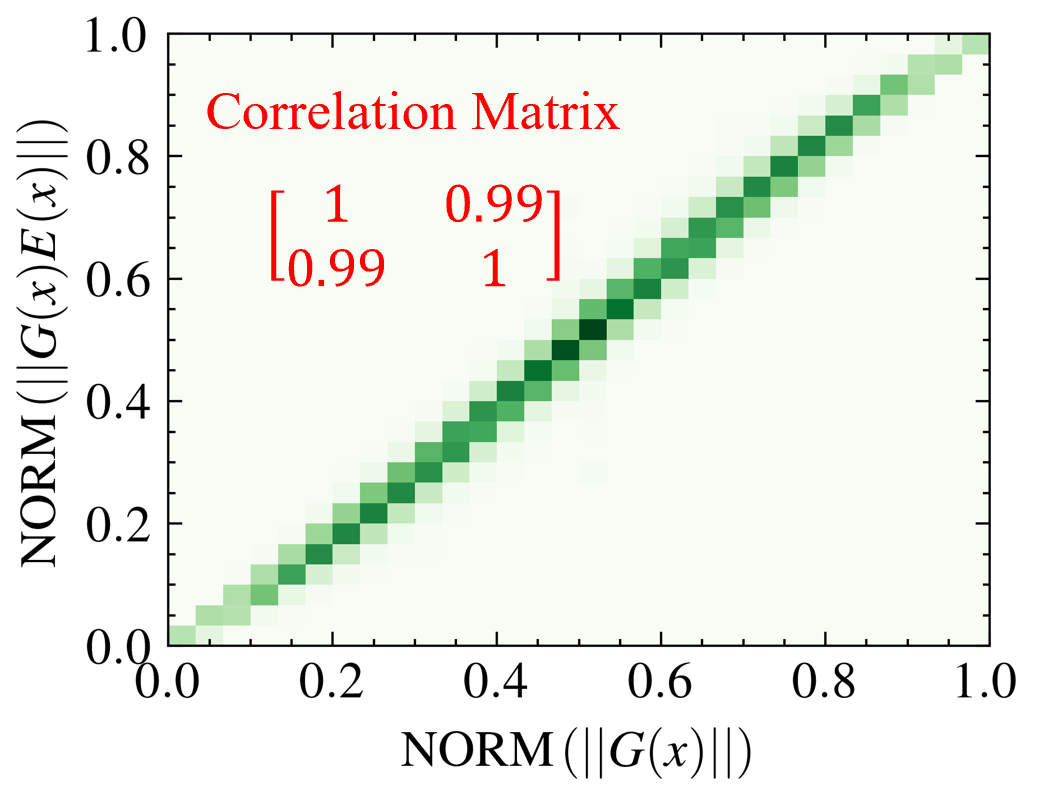}}
    \subfloat[\centering Distribution of experts' unimportance degree score]{\includegraphics[width=0.49\linewidth]{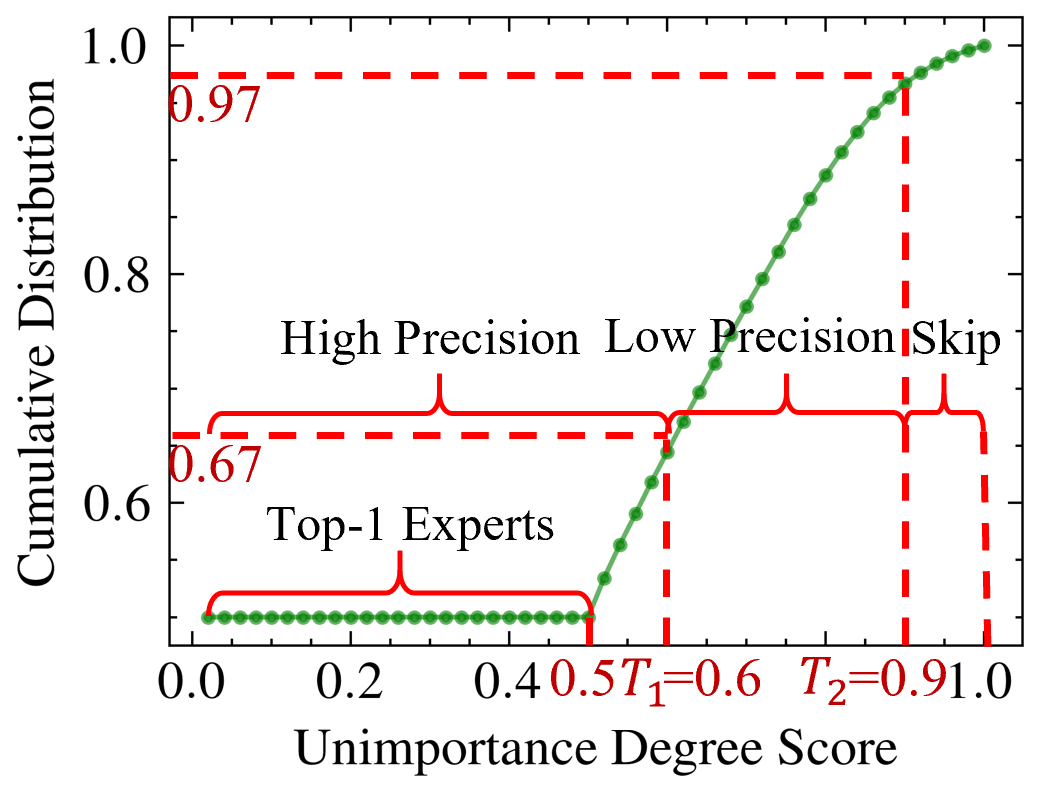}}
    \caption{Gating output statistics of Mixtral-8x7B.}
  \label{fig:gate-relation}
\end{figure}

Loading low-precision experts during cache misses effectively mitigates expert loading latency, as demonstrated in Section~\ref{sec:moti}. 
However, to preserve model accuracy, this replacement should target only less important experts. 
While model profiling on specific datasets can identify expert importance, this static approach is impractical for diverse deployment environments. Instead, we need a dynamic method to assess expert importance based on runtime inputs during the LLM's generation process.

\textbf{Expert importance estimation.} 
Based on the computing pattern of the MoE module in Equation~(\ref{eq:moe-compute}), expert \( e_i \) contributes \( G(x)_{e_i}E_{e_i}(x) \) to the output \( y \). We can represent the influence of expert \( e_i \) on the output using the magnitude \( ||G(x)_{e_i}E_{e_i}(x)|| \) (where \( ||\cdot|| \) denotes magnitude), as a smaller magnitude implies that the values in the tensor are closer to zero. Since \( E_{e_i}(x) \) cannot be computed without the weight of expert \( e_i \), we approximate \( ||G(x)_{e_i}E_{e_i}(x)|| \) using \( ||G(x)_{e_i}|| \). This approximation is based on our observation that \( ||G(x)_{e_i}|| \) and \( ||G(x)_{e_i}E_{e_i}(x)|| \) are positively correlated. To confirm this positive relationship, we collected both the expert output \( ||G(x)E(x)|| \) and the gating output \( ||G(x)|| \) from the Mixtral-8x7B model. After normalizing the data, we compute the Pearson correlation coefficient matrix and plot a heatmap to visualize their relationship. As shown in Figure~\ref{fig:gate-relation}-(a), the two variables exhibit a strong positive correlation, with a coefficient of 0.99.

\emph{\underline{Takeaways:}  
We can leverage \( ||G(x)|| \) as a computationally efficient proxy for expert importance, given its strong positive correlation with \( ||G(x)E(x)|| \).
}

\begin{figure}[t]
    \centering
    \includegraphics[width=\linewidth]{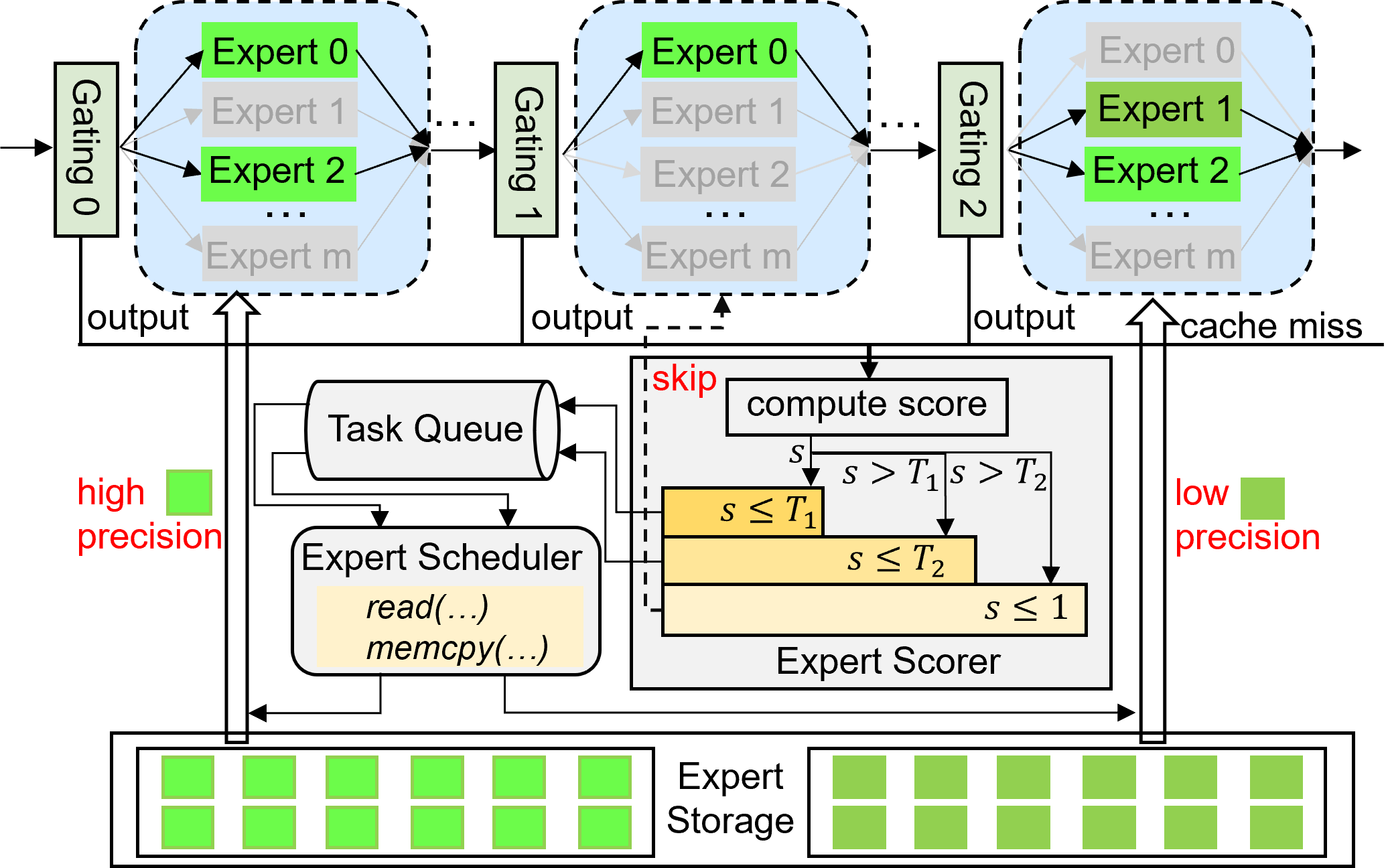}
    \caption{Token-level dynamic Expert Loader.}
    \label{fig:handler}
\end{figure}

\textbf{Expert loader design.}
Based on the observations above, we first rank the selected \( K \) experts in descending order of \( ||G(x)_{e_i}|| \) (where a larger \( i \) corresponds to a smaller \( ||G(x)_{e_i}|| \), and \( ||G(x)|| \) values are normalized). Next, we calculate the unimportance degree score \( s_{e_i} \) for each expert \( e_i \) as follows:

\begin{equation}
s_{e_i}(x) = \begin{cases}
    \sum_{j=0}^{i-1} ||G(x)_{e_j}||, & i>0 \\
    0, & i=0
\end{cases}
\end{equation}

Where $x$ is the gating input of current processing token. Thus, each expert \( e_i \) has a score to represent its importance (a higher score indicates lower importance). This score will determine whether the expert is replaced with a low-precision version. Specifically, we set a threshold \( T_1 \) (where \( 0 \leq T_1 \leq 1 \)): if \( s_{e_i} \leq T_1 \), we consider the expert important and load the high-precision version; otherwise, we opt for the low-precision version to reduce loading overhead due to its minimal influence on the output. Notably, we always treat the first expert ($e_0$) as important, keeping it in high precision to maintain model accuracy.

Based on the unimportance degree score, we implement the Dynamic Expert Loader as illustrated in Figure~\ref{fig:handler}. 
To increase flexibility, we introduce a second threshold \( T_{2} \), allowing the system to bypass less important experts. As shown in Figure~\ref{fig:handler}, when a cache miss occurs, the Expert Scorer module computes the scores of the missed experts and generates appropriate tasks based on these scores, adding them to the Task Queue. The Expert Scheduler then fetches tasks from the queue and loads the corresponding precision experts from expert storage via system calls, such as \textit{read(...)}. For instance, in the figure, Gating 0 retrieves a high-precision expert due to its high importance, Gating 1 skips an expert deemed of very low importance, and Gating 2 fetches a low-precision expert for moderate importance. To select the threshold values, we can profile the score distribution of all experts. As depicted in Figure~\ref{fig:gate-relation}-(b), we set \( T_1 = 0.6 \) and \( T_2 = 0.9 \) for the Mixtral-8x7B model, dividing the experts into three groups: 67\% in high precision, 30\% in low precision, and 3\% to skip. This configuration maintains model accuracy while significantly reducing expert-loading costs. Due to Mixtral-8x7B's top-2 selection mechanism, all top-1 experts (50\% of selections) receive scores of 0, ensuring they remain in the high-precision group.

With this method, \modelname can dynamically load experts with the appropriate precision based on the current input when a cache miss occurs, significantly reducing expert-loading latency while maintaining both model accuracy and deployment flexibility.

\subsection{Layer-level Adaptive Expert Prefetching}\label{sec:predictor-design}

\begin{figure}[t]
    \centering
    \captionsetup[subfloat]{skip=-2pt}
    \subfloat[Cosine similarity for next layers]{\includegraphics[width=1.0\linewidth]{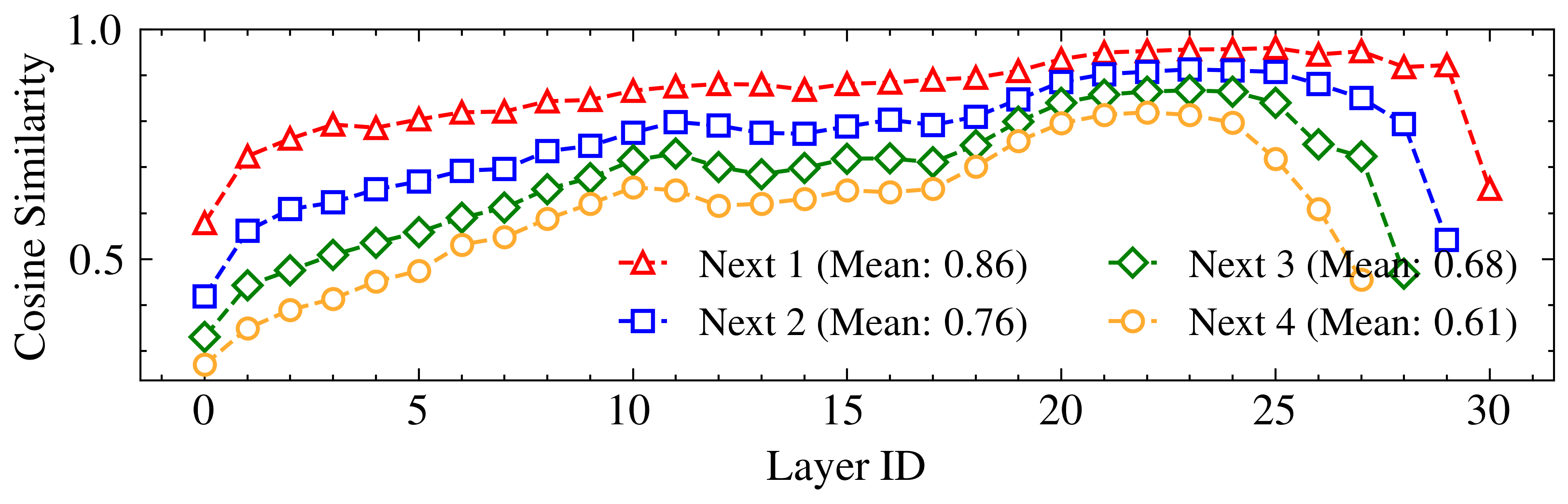}}
    \hfill
    \subfloat[Predicting accuracy for next layers]{\includegraphics[width=1.0\linewidth]{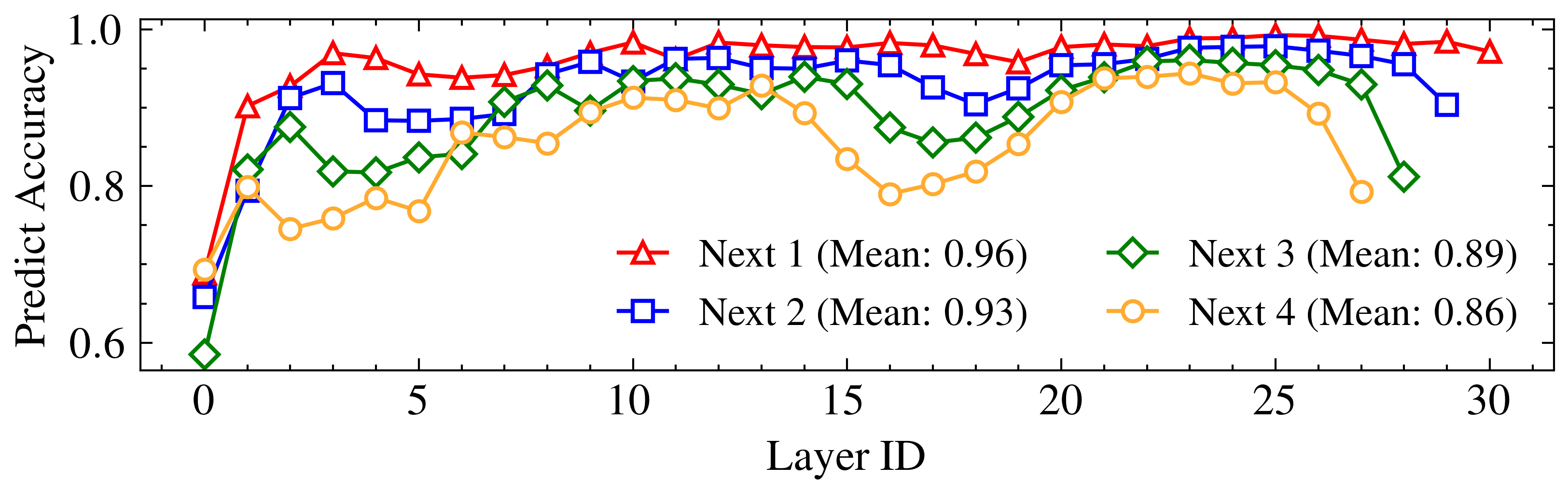}}
    \caption{Cosine similarity and predicting accuracy across layers of Mixtral-8x7B, where "Next i" refers to the next $i$-th layer from the current layer.}
  \label{fig:predict-acc-simi}
\end{figure}

To fully leverage the benefits of overlapping communication with computation, we require a highly accurate method for prefetching mixed precision experts for subsequent layers, while minimizing penalties from incorrect predictions. Due to the layer-by-layer structure of LLMs, we can explore the similarities between model layers to design the method.

\textbf{Similarity between layers.}
Due to the residual structure in LLMs, hidden states across consecutive layers exhibit significant similarity~\cite{men2024shortgpt, chen2024compressing, kim2024shortened}. This suggests that the inputs to the gating function in the MoE module also share high similarity across successive layers. As shown in Figure~\ref{fig:predict-acc-simi}-(a), the cosine similarity of gating inputs between two consecutive layers (labeled as "Next 1" in the figure) is notably high in the Mixtral-8x7B model. In fact, even the inputs for the next two and three layers exhibit considerable similarity. As a result, we can leverage the gating input from the current layer to predict the required experts for subsequent layers. Figure~\ref{fig:predict-acc-simi}-(b) demonstrates that the top-1 expert prediction accuracy for the next layer is very high, averaging 96\% across layers. Even for the next two or three layers, the accuracy remains around 90\% on average across all layers.

\emph{\underline{Takeaways:}  
We can exploit the strong layer-wise similarity of gating inputs to design an accurate and efficient expert prefetching mechanism.
}

\begin{figure}[t]
    \centering
    \includegraphics[width=\linewidth]{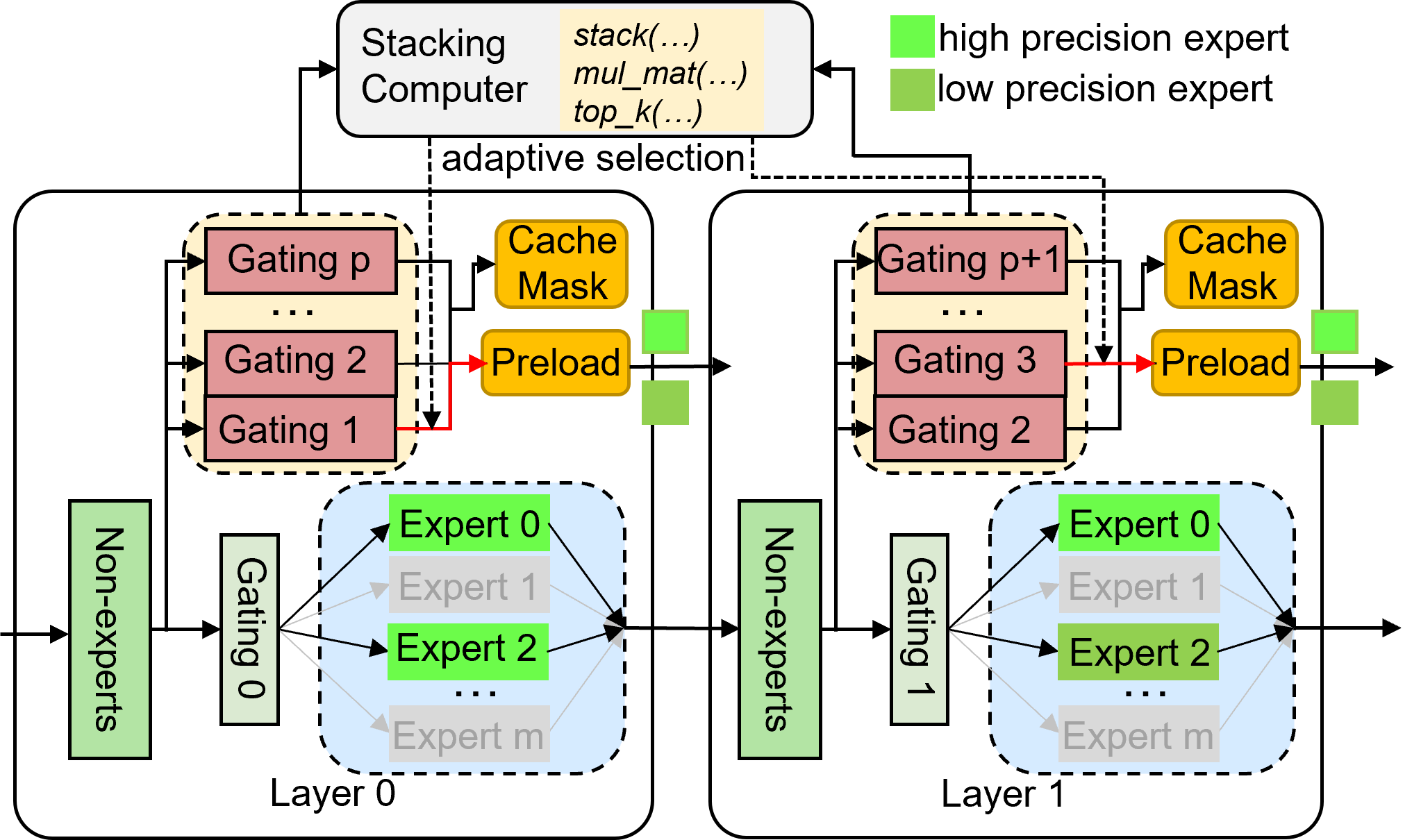}
    \caption{Layer-level adaptive Expert Predictor.}
    \label{fig:predictor}
\end{figure}

\textbf{Expert predictor design.}
Based on these observations, we build the layer-level  Adaptive Expert Predictor. As depicted in Figure~\ref{fig:predictor}, we begin by predicting the experts required for the next layer. If all predicted experts are present in the expert cache, we then proceed to predict for the subsequent layer. This process continues until either some predicted experts are missing from the cache or all predictions are completed (\(p\) gating modules per layer). For example, in layer 0 of the figure, the experts for layer 1 (gating 1) need to be preloaded, while those for layer 3 (gating 3) are required at layer 1 since the experts for layer 2 are already in the expert cache.  Furthermore, we will mask all predicted experts to prevent them from being evicted from the expert cache, as they are highly likely to be used in the subsequent  layers. And we preload versions of the experts with different precision levels to facilitate faster loading and minimize prediction penalties.

When integrating the predictor into the system, we must consider both the computational overhead of the predictor and the penalties associated with incorrect predictions. In a naive approach, the gating function would be computed sequentially until the required experts are identified, resulting in an overhead that grows linearly with the number of gating computations. Obviously, this method is inefficient. Given that one dimension of the gating module's weight corresponds to the number of experts (typically small values such as 8, 16, or 64), we can optimize the process by stacking all \( p \) gating modules together and computing them simultaneously. This approach nearly matches the computational speed of a single gating module, taking advantage of the high parallel performance offered by GPUs. Therefore, we design the Stacking Computer module to compute all \( p \) gating modules at once using several tensor operations, including \textit{stacking}, \textit{matrix multiplication}, and \textit{top-k selection}, and to adaptively select the required experts for preloading. This stacking module efficiently identifies the required experts while minimizing the overhead associated with the prefetching technique.

\begin{figure}[t]
    \centering
    \includegraphics[width=\linewidth]{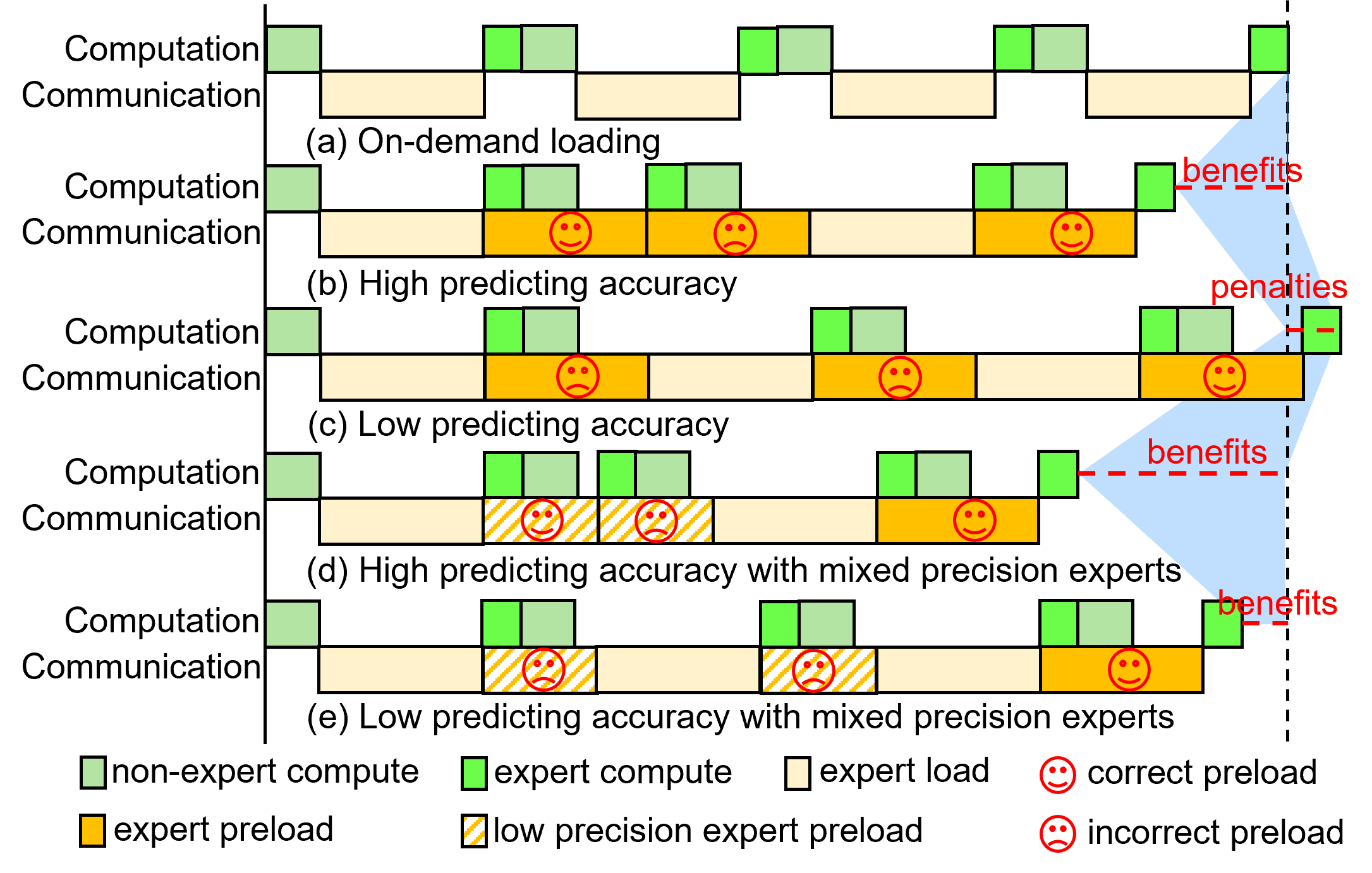}
    \caption{Preload timeline under different conditions.}
    \label{fig:prefetch}
\end{figure}

Under ideal conditions, there would be no penalties associated with incorrect predictions from the predictor, as we can halt the memory copy operation upon detecting an error and immediately initiate the loading of the correct expert. However, in practical implementations, such as with \textit{cudaMemcpy()}, we cannot interrupt the memory copy operation until it completes. As a result, we must wait for the loading of the incorrect expert to finish before we can begin loading the required experts. Given the lengthy expert-loading latency, this can lead to significant penalties when prediction accuracy is low, potentially resulting in worse performance than on-demand loading without predictions. As shown in Figure~\ref{fig:prefetch}, we can gain some benefits when prediction accuracy is high (Figure~\ref{fig:prefetch}-(b)). In contrast, low prediction accuracy results in  penalties (Figure~\ref{fig:prefetch}-(c)) due to the time costs associated with loading incorrect experts. However, our approach  leverages mixed precision expert loading mechanism to mitigate this issue. Comparing Figure~\ref{fig:prefetch}-(c) with Figure~\ref{fig:prefetch}-(e), we see that even with low prediction accuracy, we can still obtain benefits from using mixed precision expert loading, as it incurs much lower penalties than the original method. Moreover, when prediction accuracy is high, we observe  greater benefits (Figure~\ref{fig:prefetch}-(d)). 

Therefore, with the mixed precision expert loading method, \modelname can fully exploit the benefits of prefetching, achieving high prediction accuracy with minimal penalties from incorrect prefetching.

\subsection{Sequence-level Multidimensional Expert Caching}

To fully leverage the potential of the mixed precision expert cache, it is crucial to design an effective cache replacement policy that accounts for the varying loading costs of low-precision and high-precision experts.

\begin{figure}[t]
    \centering
    \subfloat[Probability of experts used between two consecutive tokens]{\includegraphics[width=1.0\linewidth]{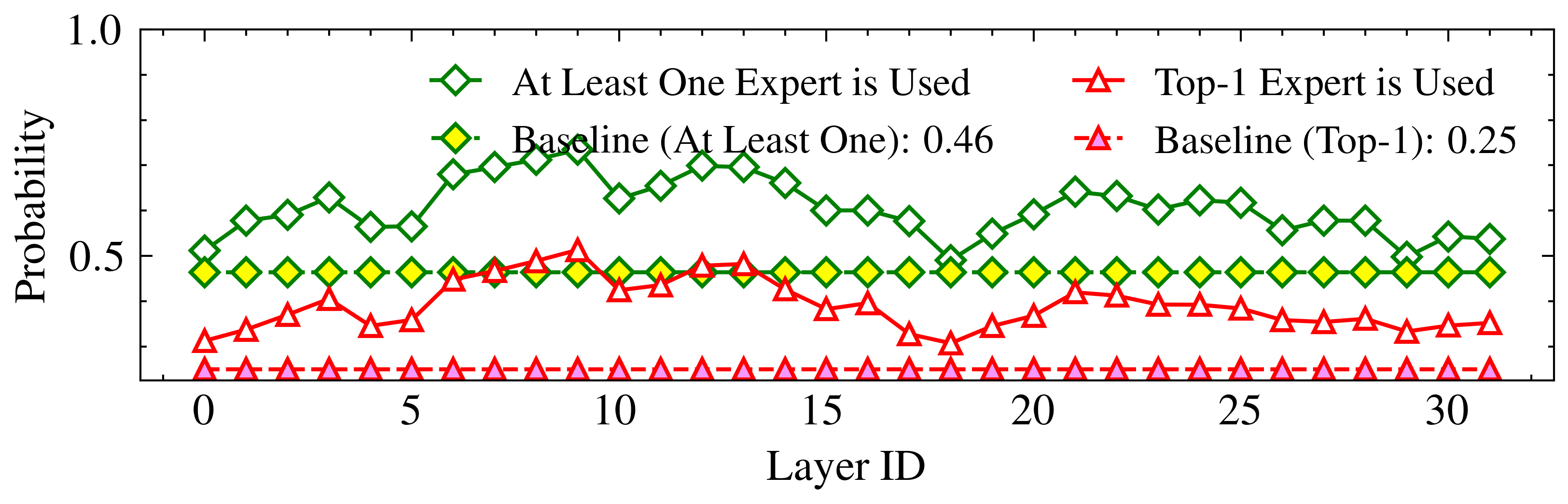}}
    \hfill
    \subfloat[Frequency of experts used in different sequences]{\includegraphics[width=1.0\linewidth]{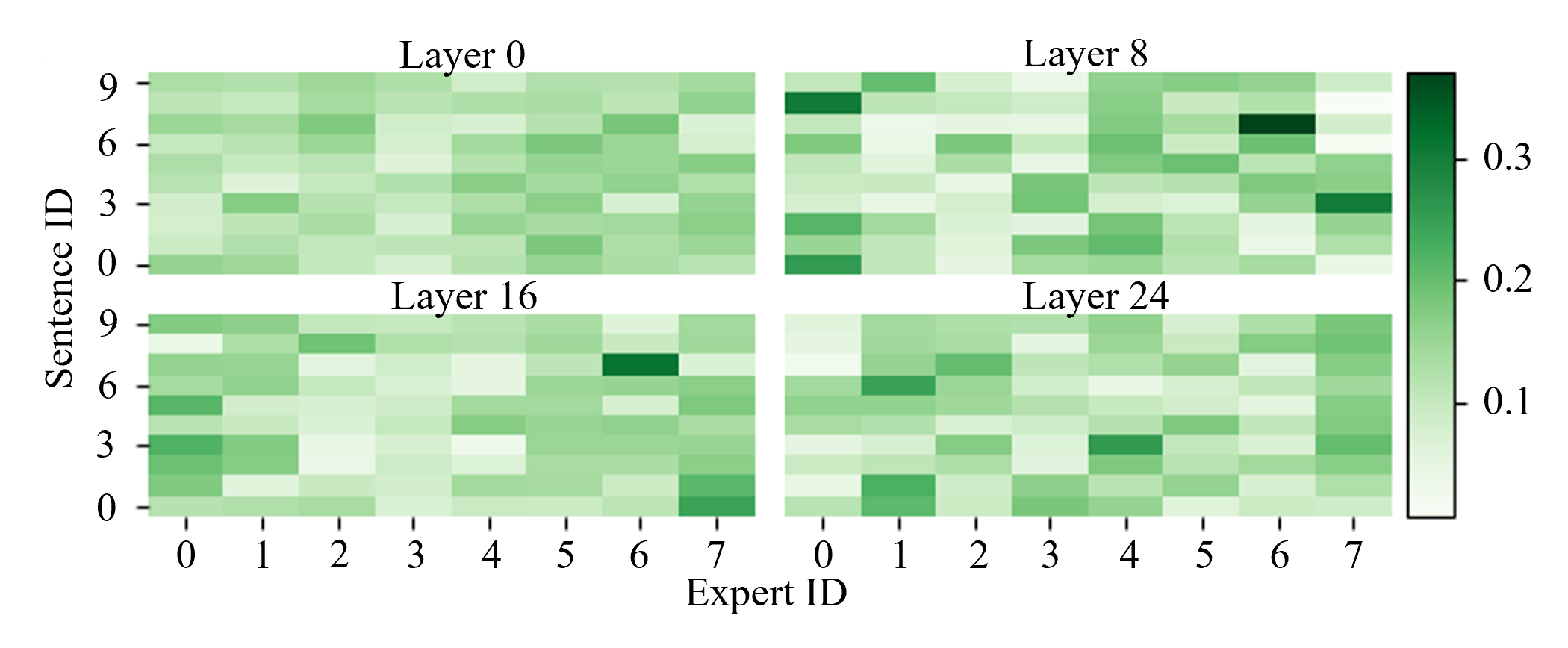}}
    \caption{Statistics of experts usage  for Mixtral-8x7B.}
  \label{fig:cache-profile}
\end{figure}

\textbf{Cache replacement policies.}
Traditionally, Least Recently Used (LRU) and Least Frequently Used (LFU) methods have been employed for cache management. Previous studies~\cite{jiang2024mixtral, eliseev2023fast} suggest that if an expert is used in the current token’s forward pass, it has a higher probability of being utilized in the next token’s forward pass, a behavior characteristic of LRU. As illustrated in Figure~\ref{fig:cache-profile}-(a), in the Mixtral-8x7B model, the top-1 expert used in the current token process has a significantly higher likelihood of being used in the next token process than the theoretical probability of 0.25, given that 2 experts are selected from a pool of 8. Moreover, the probability that at least one of the two experts used in the current token process will be reused in the next token exceeds the theoretical value of 0.46 for this model. Therefore, LRU presents a viable option.

While MoE models are typically trained with an auxiliary loss to promote uniform expert selection, the frequency of expert selection varies at the sequence level. Figure~\ref{fig:cache-profile}-(b) shows that different sequences exhibit preferences for specific experts in different layers. Therefore, a sequence-level LFU can also be a possible option. Furthermore, due to the layer-wise structure of these models, experts from nearer layers are more likely to be used, which we refer to as the Farthest Layer Distance (FLD) policy. Naturally, FLD also impacts cache performance.

For our special mixed precision expert cache, it is necessary to define a specialized cache miss penalty rather than relying solely on the cache miss ratio to evaluate replacement policies, as experts of different precisions incur different penalties. Specifically, if an expert is missed, the cost of loading its high-precision version is $C$, while the cost for the low-precision version is $\frac{B_{l}}{B_{h}} C$, where $B_{l}$ and $B_{h}$ are the bit-widths of the low- and high-precision versions, respectively. For instance, if an int4 precision expert is used as a replacement for a float16 precision expert, missing the high-precision expert would incur four times the cost compared to missing the low-precision expert. Therefore, a new policy is required to handle the mixed precision scenario and minimize the miss penalties effectively.

\emph{\underline{Takeaways:}  
Model-specific behaviors and mixed precision characteristics necessitate a comprehensive caching policy that integrates multiple replacement strategies.
}

\begin{figure}[t]
    \centering
    \captionsetup[subfloat]{skip=-1pt}
    \subfloat[Frequency of mixed precision expert usage]{\includegraphics[width=\linewidth]{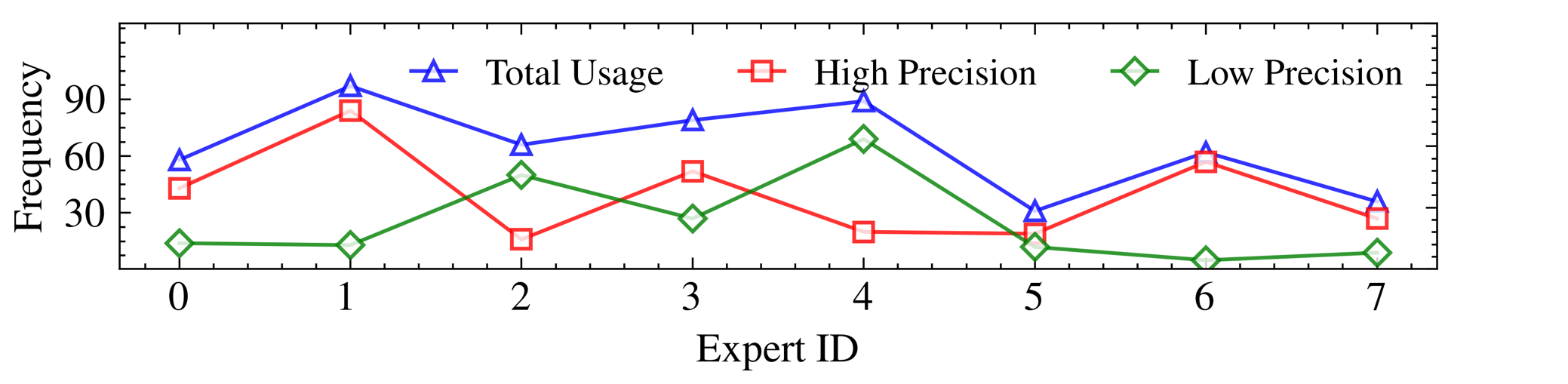}}
    \hfill
    \subfloat[Behaviors of LFU and LHU]{\includegraphics[width=\linewidth]{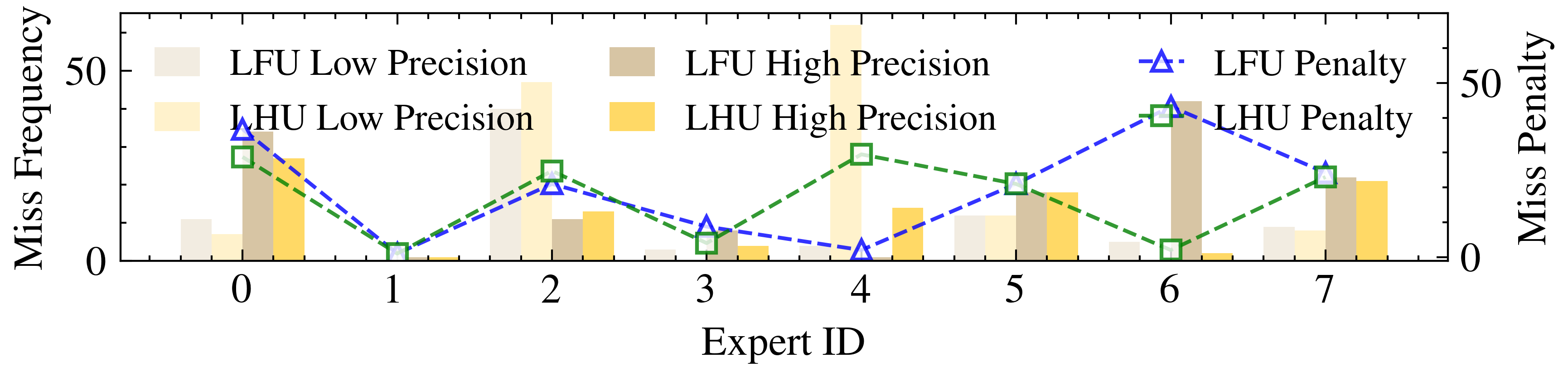}}
    \caption{Statistics of mixed precision expert usage in one layer of Mixtral-8x7B.}
  \label{fig:lhu-profile}
\end{figure}

\textbf{Cache Manager design.}
To minimize cache miss penalties in the mixed precision expert cache, we propose the Least High Precision Frequently Used (LHU) method, which prioritizes frequently used high-precision experts, as cache misses for these experts incur significantly higher latency compared to low-precision ones. Similar to LFU, we track the number of times each expert’s high-precision version is used. As shown in Figure~\ref{fig:lhu-profile}-(a), the frequency of high-precision usage is not always the same as the total frequency of usage. For example, expert 4 has a high total usage frequency but a low high-precision frequency, while expert 6 has a higher high-precision frequency despite a lower total usage frequency compared to expert 4. Consequently, LFU would prioritize expert 4, while LHU would prioritize expert 6, making LHU a distinct policy from LFU in this context.

Figure~\ref{fig:lhu-profile}-(b) shows the performance comparison between LFU and LHU for these experts. The results indicate that LHU causes more cache misses for expert 4, especially for its low-precision version, while LFU keeps expert 4 in the cache with fewer misses. However, for expert 6, LHU gives higher priority and results in fewer misses, especially for the high-precision version. Since expert 6 relies more on high-precision versions, LHU reduces cache miss penalties more effectively than LFU (here, suppose a high-precision miss incurs a penalty of 1, while a low-precision miss incurs a penalty of \(\frac{1}{4}\)), even though it performs worse for expert 4.  A similar pattern can be observed with expert 0 and expert 2. Overall, for these experts, LHU reduces cache miss penalties by about 15\% compared to LFU totally. Therefore, LHU is a more suitable policy in our specific scenario than LFU.

\begin{figure}[t]
    \centering
    \includegraphics[width=\linewidth]{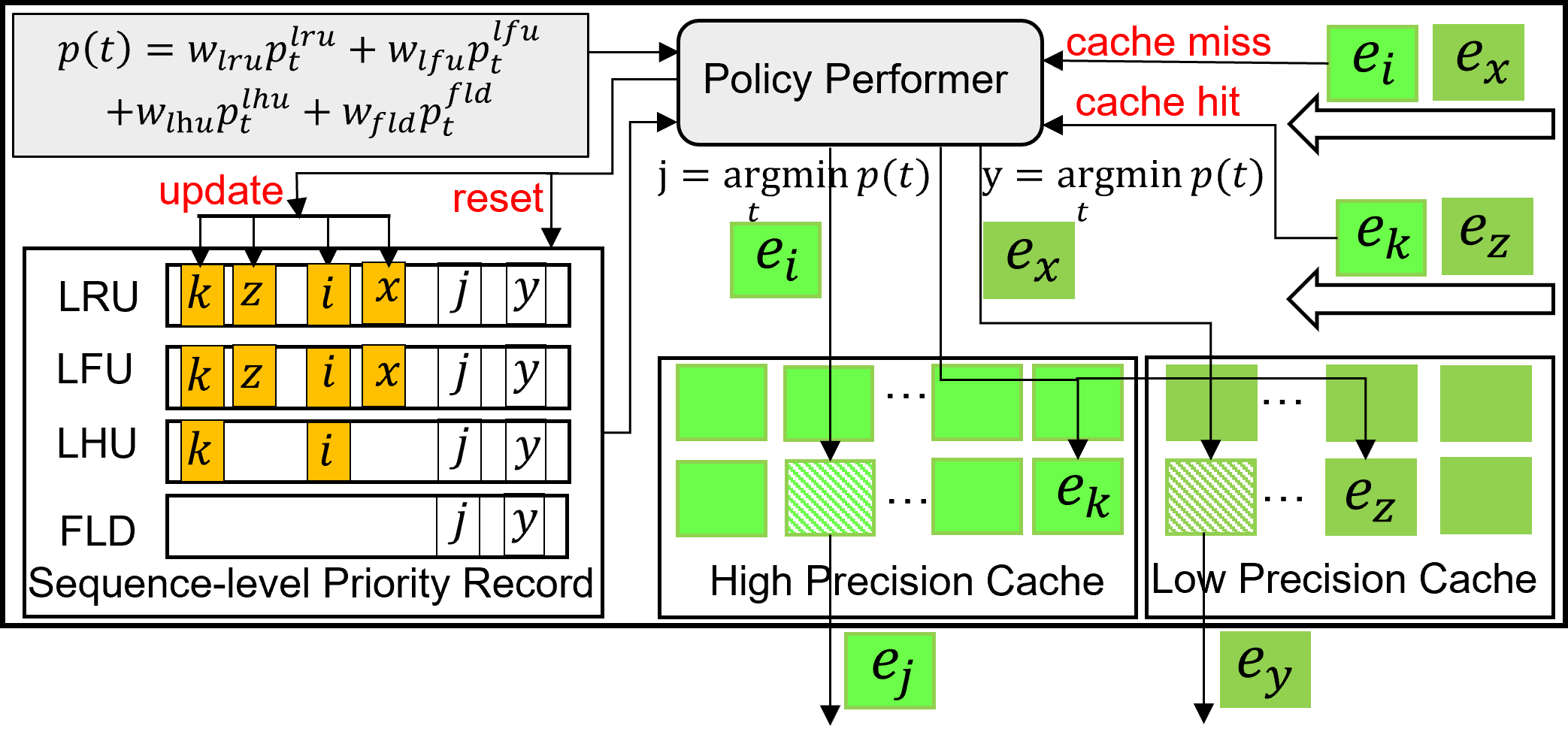}
    \caption{Sequence-level multidimensional Cache Manager.}
    \label{fig:expert-cache}
\end{figure}

To leverage the strengths of the above four policies, we combine them by assigning weights to each and summing them to determine the final priority for each expert. We define the priority of an expert \( t \) as follows:

\begin{equation}\label{eq:cache-policy}
\begin{cases}
      & p_t = w_{\text{lru}}p_{t}^{\text{lru}} + w_{\text{lfu}}p_{t}^{\text{lfu}} + w_{\text{lhu}}p_{t}^{\text{lhu}} +  w_{\text{fld}}p_{t}^{\text{fld}}  \\
      & w_{\text{lru}} + w_{\text{lfu}}  + w_{\text{lhu}} + w_{\text{fld}} = 1 \\
      & p_{t}^{\text{lru}} = \frac{R_t}{T};
       p_{t}^{\text{lfu}} = \frac{F_t}{T};
       p_{t}^{\text{lhu}} = \frac{H_t}{T};
       p_{t}^{\text{fld}} = 1 - \frac{(l_t - l_i + l_n) \% l_n}{l_n}
\end{cases}
\end{equation}
Where \( R_t \) represents the last-used time of expert \( t \), \( F_t \) denotes the frequency of use in the current processing sequence, \( H_t \) is the frequency of high-precision usage in the current sequence, \( T \) is the current token number, \( l_i \) is the layer ID of the currently used expert, \( l_t \) is the layer ID of expert \( t \), and \( l_n \) is the total number of layers in the model. The four weights are hyperparameters set by the user, and we determine suitable values by minimizing the mixed precision expert cache miss penalties on a calibration dataset.

Using this equation, we can identify the expert \( j \) with the lowest priority in the cache, relative to the current expert \( i \), and replace \( j \) with \( i \). As shown in Figure~\ref{fig:expert-cache}, We build the Multidimensional Cache Manager based on this equation and data recording.   The Cache Manager maintains separate caches for high- and low-precision experts, with the high-precision cache typically being larger than the low-precision cache. Whenever a high-precision expert \( e_i \) is added to the cache (a cache miss), The Policy Performer module in Cache Manager will update corresponding LRU, LFU, and LHU records and determine the lowest-priority expert \( e_j \) based on Equation~(\ref{eq:cache-policy}) and data in the priority record. The Policy Performer then evict \( e_j \) and replace it with \( e_i \) in the high-precision cache. Similarly, for a low-precision expert \( e_x \), The Policy Performer performs the same operation: update records, evict \( e_y \) from the low-precision cache, and store \( e_x \), though LHU record is not updated for low-precision experts. On a cache hit, The Policy Performer only updates the relevant records (e.g., \( e_k \) and \( e_z \) in the figure). Additionally, at the start of each new sequence, The Policy Performer resets the LRU, LFU and LHU records.

By fully leveraging the characteristics of MoE models and unique features of mixed expert cache, The Multidimensional Cache Manager can efficiently manage the cache and achieve lower cache miss penalties than previous approaches, resulting in faster inference.

\begin{figure}[t]
    \centering
    \includegraphics[width=\linewidth]{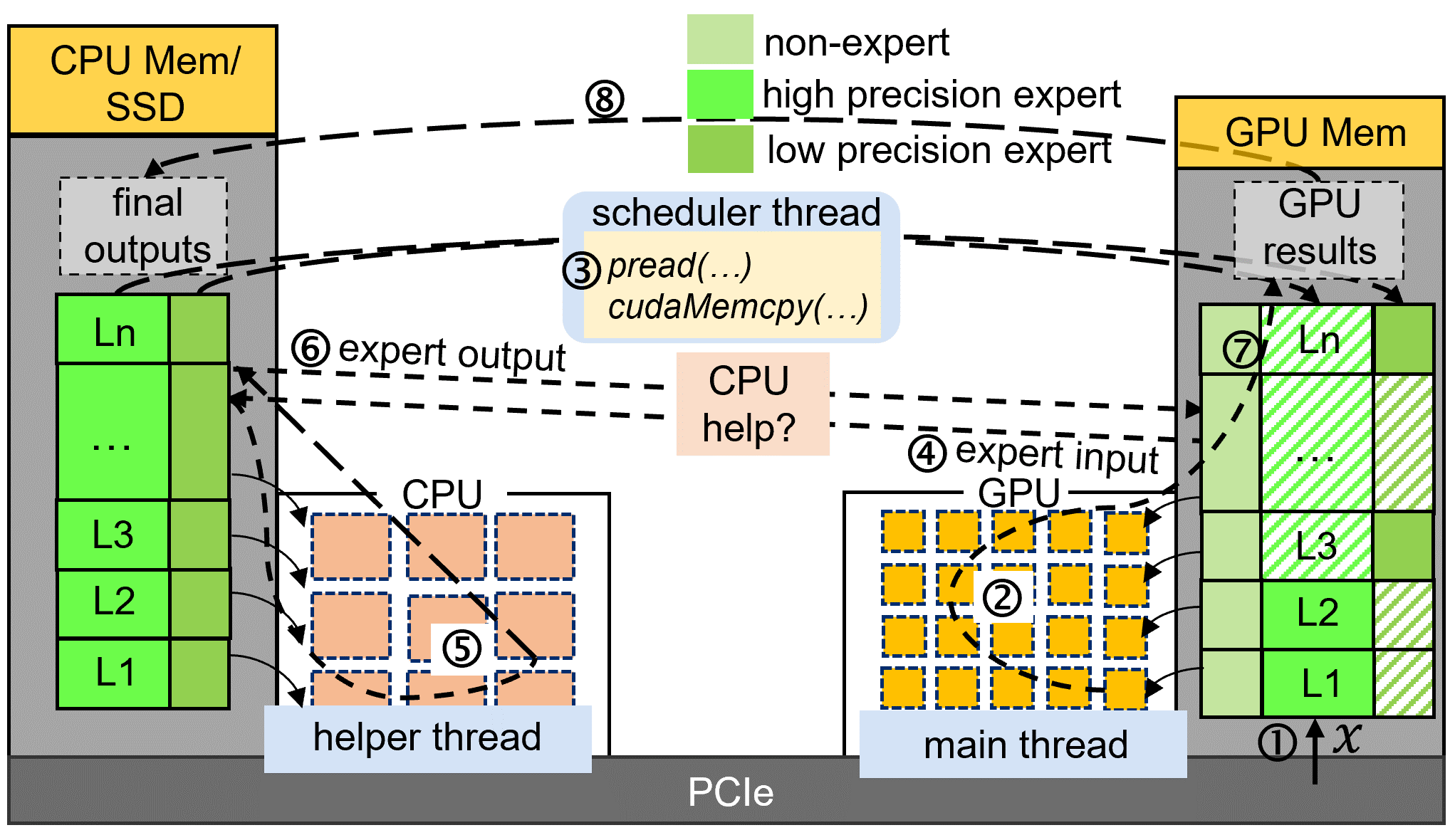}
    \caption{The implementation of \modelnamenospace.}
    \label{fig:system-workflow}
\end{figure}

\section{System Implementation}

We build our system on top of Llama.cpp by modifying the distribution of model weights and computation patterns, implemented with 8,000 lines of C++/C code. The Llama.cpp system places a sufficient number of layers in GPU memory, with the remaining layers stored in CPU memory or on SSD. It processes input on the GPU using layers in GPU memory, then sends the internal activations of the last GPU-processed layer to the CPU. It continues processing with the remaining layers on the CPU, and finally gets the results. While this computation pattern works well for dense models, it is not optimal for MoE-based LLMs.

To optimize our system for MoE models, we modify the distribution of model weights. As illustrated in Figure~\ref{fig:system-workflow}, we place all non-expert weights and a portion of experts, in multiple precision versions, in GPU memory, and all expert weights reside in CPU memory. To ensure the system performs efficiently across various hardware setups, we implement two computing modes: GPU-centric computing and CPU-GPU cooperative computing.

In the GPU-centric computing mode, when \ding{172} input $x$ is processed, the main thread \ding{173} handles it on the GPU using the corresponding model weights. If the required expert are not in GPU memory, the scheduler thread \ding{174} loads the appropriate version of the expert from CPU memory or SSD through specific interfaces. Once the required expert is loaded into GPU memory, the main thread \ding{178} resumes computation and eventually \ding{179}  transfers the final results back to the CPU.

In the CPU-GPU cooperative computing mode, if the required expert is not in GPU memory, the main thread \ding{175} sends the expert's input to the CPU, where a helper thread \ding{176} processes it using the corresponding expert. The helper thread then \ding{177} sends the expert's output back to the GPU. Once receiving the data, the main thread \ding{178} continues the computation and \ding{179} copies the results back to the CPU.

While both computing modes work well for MoE models, we primarily focus on the GPU-centric computing mode, as CPU resources are typically insufficient on edge devices.

\section{Experimental Evaluation}

\subsection{Experimental Methodoloy}

\noindent \textbf{Hardware.}
To evaluate \modelname in different environments, we use two common edge devices: the NVIDIA GeForce RTX 4090~\cite{rtx-4090} and the NVIDIA Jetson AGX Orin~\cite{jetson-orin}. In our setup, the RTX 4090 has 24GB of GPU memory, 256GB of CPU memory, and 64 CPU cores. The connection between the CPU and GPU is via PCIe 4.0, offering a theoretical bandwidth of 32GB/s. The Jetson Orin, on the other hand, has 32GB of unified memory, shared with its 12 CPU cores. For model weight storage, we use a Samsung NVMe SSD 980 PRO~\cite{980pro-ssd}, which provides a theoretical read speed of 7,000 MB/s (around 3,000 MB/s in practice).

\noindent \textbf{Models.} 
We evaluate our system using two popular MoE-based LLMs from Huggingface Hub~\cite{huggingfacehub}: Mixtral-8x7B~\cite{jiang2024mixtral} and Phi-MoE~\cite{abdin2024phi}. As shown in Table~\ref{tab:moe-model}, Mixtral-8x7B has 45 billion parameters across 32 layers, with 8 experts per layer. For each token, 2 experts are selected per layer, activating 14 billion parameters during a forward pass. Phi-MoE, on the other hand, has 42 billion parameters across 32 layers, with 16 experts per layer, activating 6.6 billion parameters per forward pass.

\begin{table}[t]
\centering
\small
\renewcommand\arraystretch{1.0}
\caption{Configuration of two evaluated MoE models.}
\resizebox{\linewidth}{!}{
\begin{tabular}{lcc}
\toprule
                 & \textbf{Mixtral-8x7B}  & \textbf{Phi-MoE}    \\ 
\midrule
Total Parameters       & 45B  & 42B                 \\ 
Actived Parameters/Token & 14B & 6.6B               \\ 
Total Weight Size         & 87GB  & 78GB             \\ 
Experts  Weight Size       & 84GB(96\%) & 75GB(96\%)     \\ 
Layer Number      & 32        & 32           \\ 
Expert Number/Layer    & 8   & 16                  \\ 
Top-K              & 2         & 2            \\ 
\bottomrule
\end{tabular}
}
\label{tab:moe-model}
\end{table}

\noindent \textbf{Datasets.}
To test the generation speed of models, we extract 60 high-quality samples from the Alpaca~\cite{alpaca} dataset. Half of these samples have an input length of 16, while the other half have an input length of 128. To evaluate the impact of \modelname on model accuracy, we use GSM8K~\cite{cobbe2021gsm8k} and TruthfulQA~\cite{lin2021truthfulqa} as performance evaluation datasets. GSM8K is designed to assess the model’s mathematical reasoning abilities, while TruthfulQA is used to measure whether a language model generates truthful answers.

\noindent \textbf{Baselines.}
We compare \modelname (\modelnameshortnospace) with six SOTA inference systems to evaluate its efficiency. (1) Transformers~\cite{wolf-etal-2020-transformers} (TF), A general LLM library developed by Huggingface, offering thousands of pretrained models. (2) DeepSpeed-Inference~\cite{deepspeedinference} (DS), A comprehensive inference system for LLMs, providing multi-GPU and heterogeneous inference solutions. (3) Llama.cpp~\cite{llamacpp} (LL), An efficient LLM inference system written in pure C/C++, supporting simultaneous computation on both CPU and GPU. (4) MoE-Offloading~\cite{eliseev2023fast} (MO), A MoE-centric system that incorporates expert prediction and caching. (5) MoE-Infinity~\cite{xue2024moe} (MI), A system that tracks request-level processes to prefetch required experts into GPU memory. (6) Fiddler~\cite{kamahori2024fiddler} (FD), a system that leverages CPU computation to minimize data movement between the CPU and GPU. Additionally, since some systems do not natively support the Phi-MoE model, we integrated it into them following their respective guidelines.

\begin{table}[t]
\centering
\renewcommand\arraystretch{1.0}
\caption{Configuration of three testing groups.}
\resizebox{\linewidth}{!}{
\begin{tabular}{ccc}
\toprule
     \textbf{Hardware Setup}         & \textbf{Model Version}  &  \textbf{Methods}  \\ 
\midrule
  Jetson AGX Orin       &  Int8  &\modelnameshortnospace, LL, MI              \\ 
\midrule
  GeForce RTX 4090 & Float16 & \modelnameshortnospace, TF, DS, MO, MI \\ 
\midrule
RTX 4090 + CPU  & Float16  &\modelnameshortnospace, LL, FD        \\ 
\bottomrule
\end{tabular}
}
\label{tab:compare-settings}
\end{table}

\noindent \textbf{Configurations.}
Due to platform differences, we use different configurations to evaluate baselines. On the RTX 4090, we employ two models with float16 precision. Since Llama.cpp and Fiddler utilize CPU computation, which follows a different computational pattern from other methods, we compare them separately for fairness. On the Jetson Orin, we use the int8 precision versions, as the float16 versions are too large and slow to run due to the SSD’s slow read speed. Additionally, we only evaluate Llama.cpp and MoE-Infinity on the Jetson Orin, as the other baselines perform poorly on this device. Furthermore, for \modelname, we use int4 precision versions as replacements for the float16 precision models and int2 versions for the int8 precision models to support dynamic precision expert loading. Overall, there are three comparison group settings, as shown in Table~\ref{tab:compare-settings}.

\begin{figure*}[t] %
    \centering
    \captionsetup[subfloat]{skip=-1pt}
    \subfloat[Mixtral-8x7B on Jetson Orin]{\includegraphics[width=0.49\textwidth]{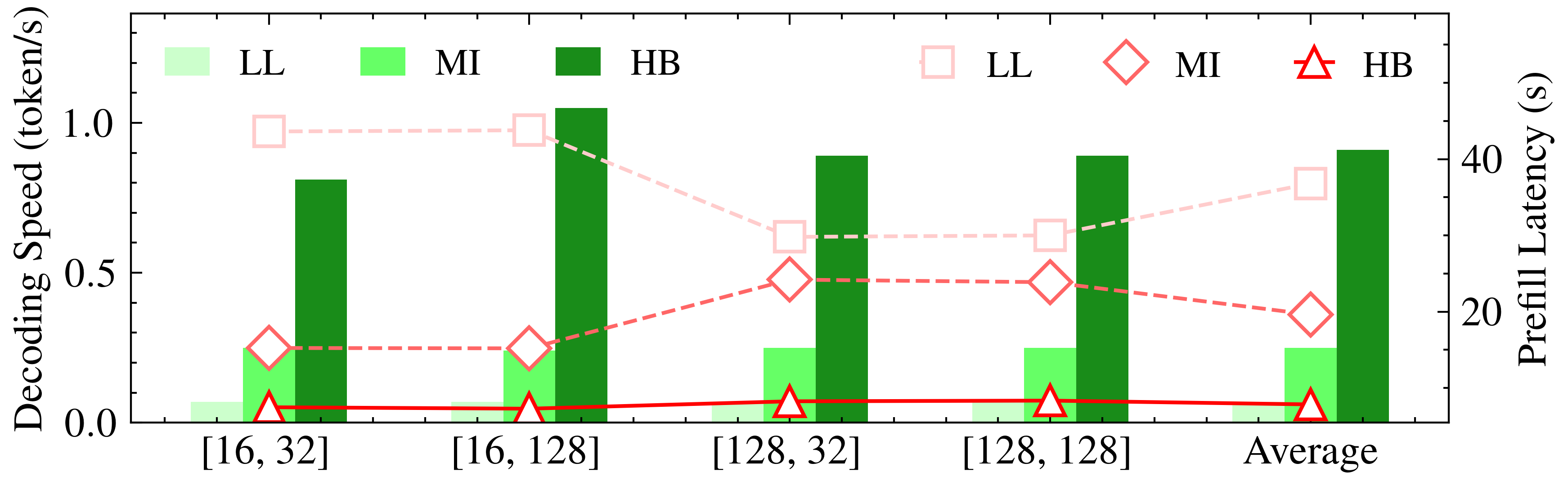} } \hfill
    \subfloat[Phi-MoE on Jetson Orin]{\includegraphics[width=0.49\textwidth]{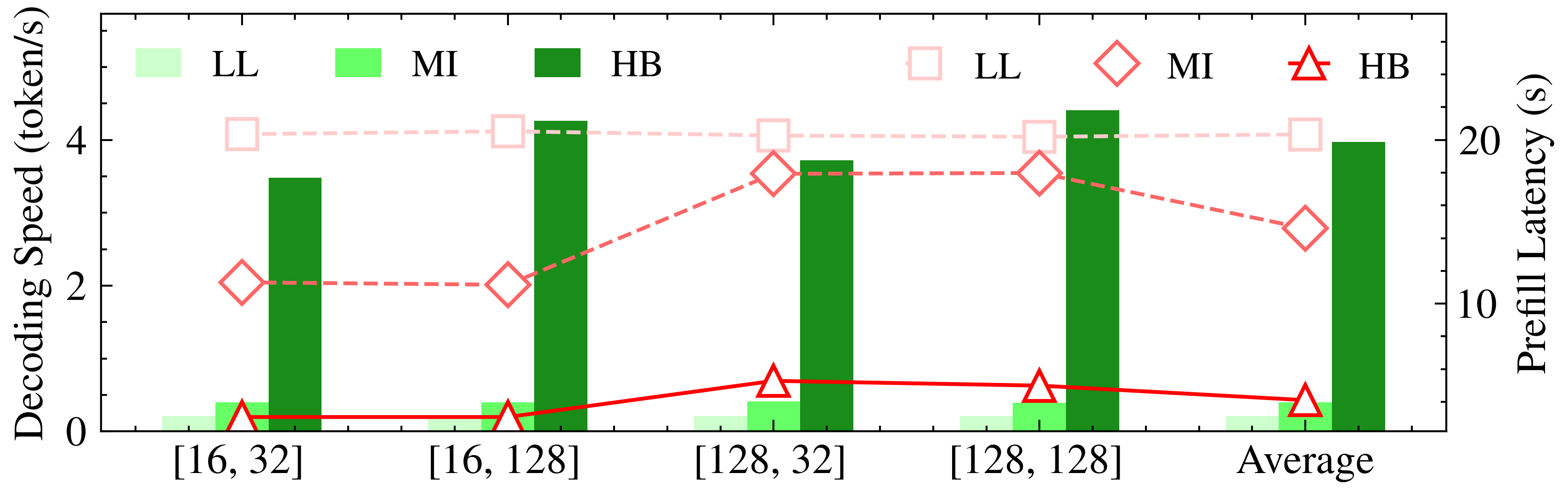} }\\
    \subfloat[Mixtral-8x7B on RTX 4090]{\includegraphics[width=0.49\textwidth]{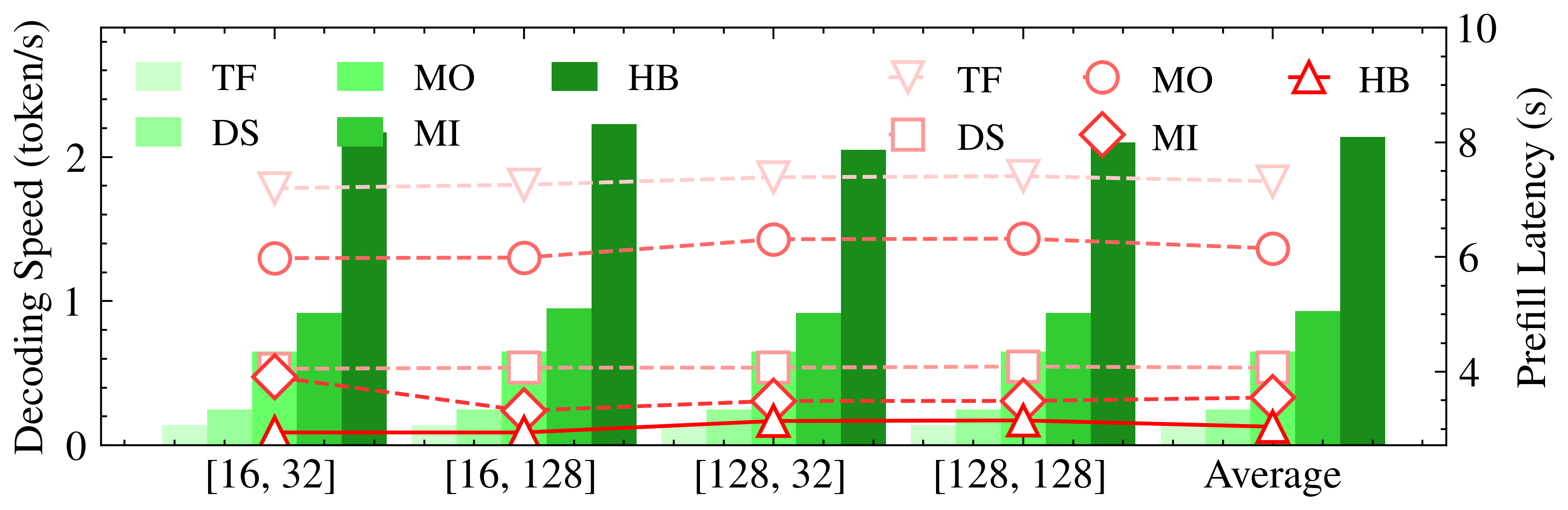} } \hfill
    \subfloat[Phi-MoE on RTX 4090]{\includegraphics[width=0.49\textwidth]{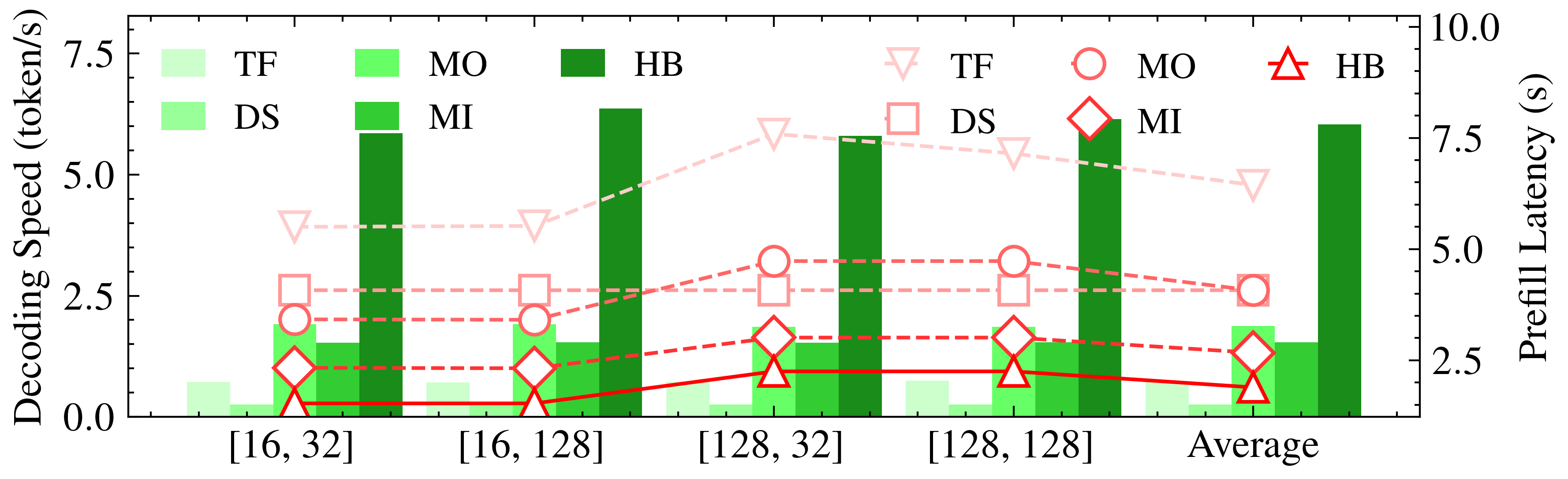} } 
    \caption{Comparison of inference speed for \modelname and the SOTA approaches.}
    \label{fig:benchmark-baselines}
\end{figure*}

\noindent \textbf{Metrics.}
Since the generation process of LLMs consists of two phases (the prefill stage and the decoding stage), we use prefill latency (in seconds) and decoding speed (in tokens per second) as our performance metrics. To strengthen and diversify the results, we set four testing groups with different input and output lengths, including [16, 32], [16, 128], [128, 32], [128,128]. And we set the batch size to 1 in all cases, following prior works~\cite{eliseev2023fast, kong-etal-2024-swapmoe, yi2023edgemoe, hwang2024pre}, as edge-side continuous serving scenarios often focus on single-batch inference.

\subsection{End-to-End Performance}

To evaluate the inference speed of \modelnamenospace, we conduct the experiments outlined in Table~\ref{tab:compare-settings} and obtain the results shown in Figure~\ref{fig:benchmark-baselines}.

From Figure~\ref{fig:benchmark-baselines}-(a) and (b), we can observe that \modelname  delivers the best performance in terms of both decoding speed and prefill latency for the evaluated models on the Jetson AGX Orin, outperforming both Llama.cpp and MoE-Infinity. Llama.cpp utilizes the \textit{mmap()} interface for fast model loading, but this can result in severe page faults when there is insufficient CPU memory to store the model weights. Since Jetson AGX Orin shares memory between the CPU and GPU, allocating memory for GPU computation leaves limited memory available for the CPU, leading to performance degradation due to frequent page faults. MoE-Infinity, primarily designed for GPU servers, also faces challenges on Jetson AGX Orin due to the slower SSD read speeds. Compared to Llama.cpp, \modelname achieves an average speedup of 13.0x for Mixtral-8x7B and 18.9x for Phi-MoE in decoding speed, along with a 79\% and 80\% reduction in prefill latency for these two models, respectively.  Although \modelname is built on Llama.cpp, it shows significantly greater advantages when deploying MoE-based LLMs on edge embedded devices. In comparison to MoE-Infinity, \modelname delivers an average speedup of 3.64x and 9.93x in decoding speed, along with a 60\% and 72\% reduction in prefill latency for these two evaluated models, respectively. These improvements are largely  attributed to its dynamic precision expert loading mechanism, which reduces the volume of data to be read, thereby accelerating inference speed.

\begin{table}[t]
\centering
\small
\caption{ Model accuracy with mixed precision experts.}
\resizebox{\linewidth}{!}{
\begin{tabular}{clccc}
\toprule
\multicolumn{2}{l}{}                                           &  \makecell{\textbf{GSM8K} \\( accuracy)} &  \makecell{\textbf{TruthfulQA} \\ (truth, info)} \\
\midrule
\multirow{2}{*}{Mixtral-8x7B } & Float16 & 0.52  & 0.49, 0.94              \\
                                      & Float16+Int4   & 0.51  & 0.49, 0.93              \\
\midrule
\multirow{2}{*}{Mixtral-8x7B}    & Int8 & 0.52  & 0.50, 0.91              \\
                                      & Int8+Int2   & 0.51  & 0.52, 0.90              \\
\midrule
\multirow{2}{*}{Phi-MoE}      & Float16 & 0.83  & 0.82, 0.92              \\
                                      & Float16+Int4    & 0.82  & 0.82, 0.93              \\
\midrule
\multirow{2}{*}{Phi-MoE}         & Int8 & 0.83  & 0.82, 0.93              \\
                                      & Int8+Int2  & 0.83  & 0.82, 0.93  \\
\bottomrule
\end{tabular}
}
\label{tab:model-acc}
\end{table}

On the GeForce RTX 4090, the results shown in Figure~\ref{fig:benchmark-baselines}-(c) and (d) demonstrate that \modelname  again outperforms the other baselines. Transformers and DeepSpeed-Inference show poor performance compared to MoE-based systems, as they do not leverage the sparse activation feature of MoE and load model parameters layer by layer on-demand. MoE-Offloading and MoE-Infinity exhibit varying performance in decoding speed depending on the model, with MoE-Infinity performing better on Mixtral-8x7B, while MoE-Offloading excels with Phi-MoE. However, MoE-Infinity consistently achieves lower prefill latency than MoE-Offloading for its good prefetching technique. Despite outperforming Transformers and DeepSpeed-Inference, both MoE-Offloading and MoE-Infinity suffer from long expert-loading latency. In contrast, compared to MoE-Offloading, \modelnamenospace, by leveraging dynamic precision expert loading, delivers an average speedup of 3.21x for Mixtral-8x7B and 3.29x for Phi-MoE in decoding speed, along with a 51\% and 54\% reduction in prefill latency, respectively. Compared to MoE-Infinity, \modelname achieves a 2.30x and 3.92x speedup in decoding speed and a 14\% and 29\% reduction in prefill latency for these two models, respectively. 
Therefore, on edge server platforms with sufficient CPU memory to store model parameters, \modelname continues to perform well for running MoE-based LLMs.

Overall, our system achieves significant speedup in decoding speed and a large reduction in prefill latency compared to the baselines across different environments, demonstrating its superior efficiency.

\subsection{Model Accuracy}

Since we use low precision experts to replace high precision experts for faster expert loading, it is crucial to verify the model's accuracy to ensure that our system does not compromise performance. As shown in Table~\ref{tab:model-acc}, we report the average \textit{top-1} accuracy for GSM8K, as well as the truthful and informative scores for TruthfulQA, following the official guidelines. After applying the mixed precision expert policy, the model accuracy decreases by no more than 1\% across both tested datasets for all evaluated models. This demonstrates that our system achieves high inference speed while maintaining model accuracy.

\subsection{CPU Computation Assistant}
\begin{figure}[t]
    \centering
    \captionsetup[subfloat]{skip=-1pt}
    \subfloat[Results of Mixtral-8x7B]{\includegraphics[width=\linewidth]{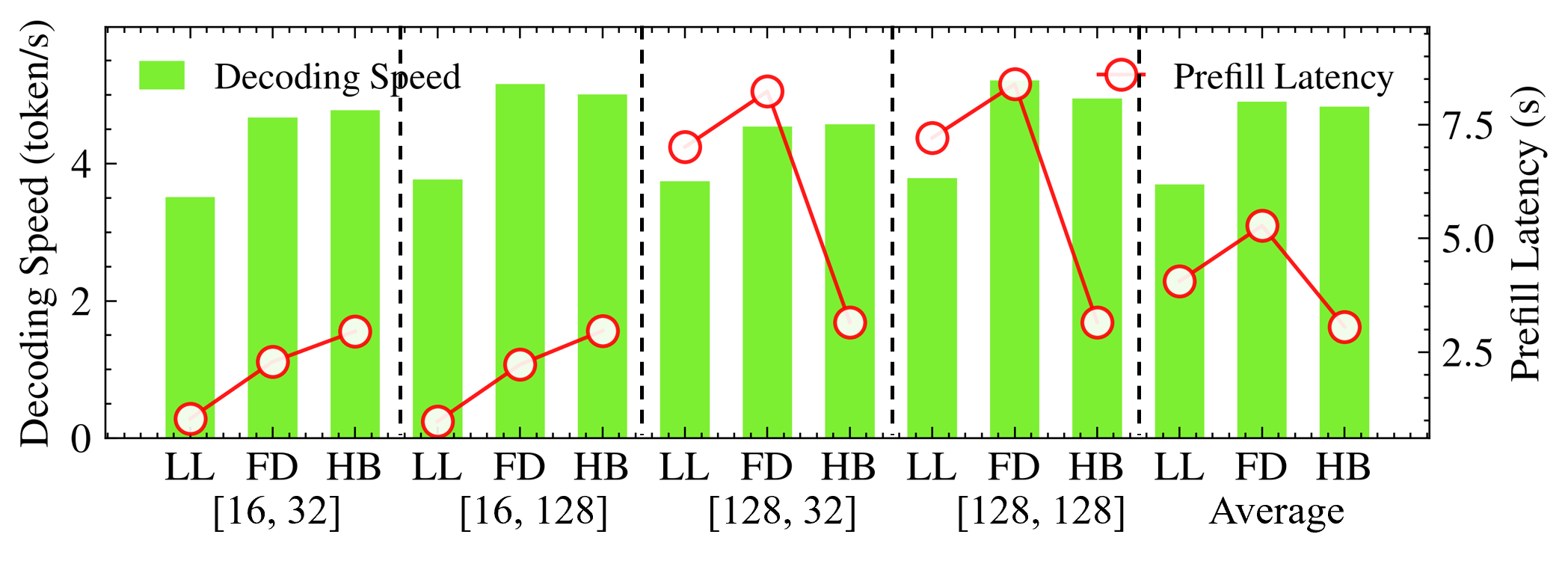}}
    \hfill

   \textit{} \subfloat[Results of Phi-MoE]{\includegraphics[width=\linewidth]{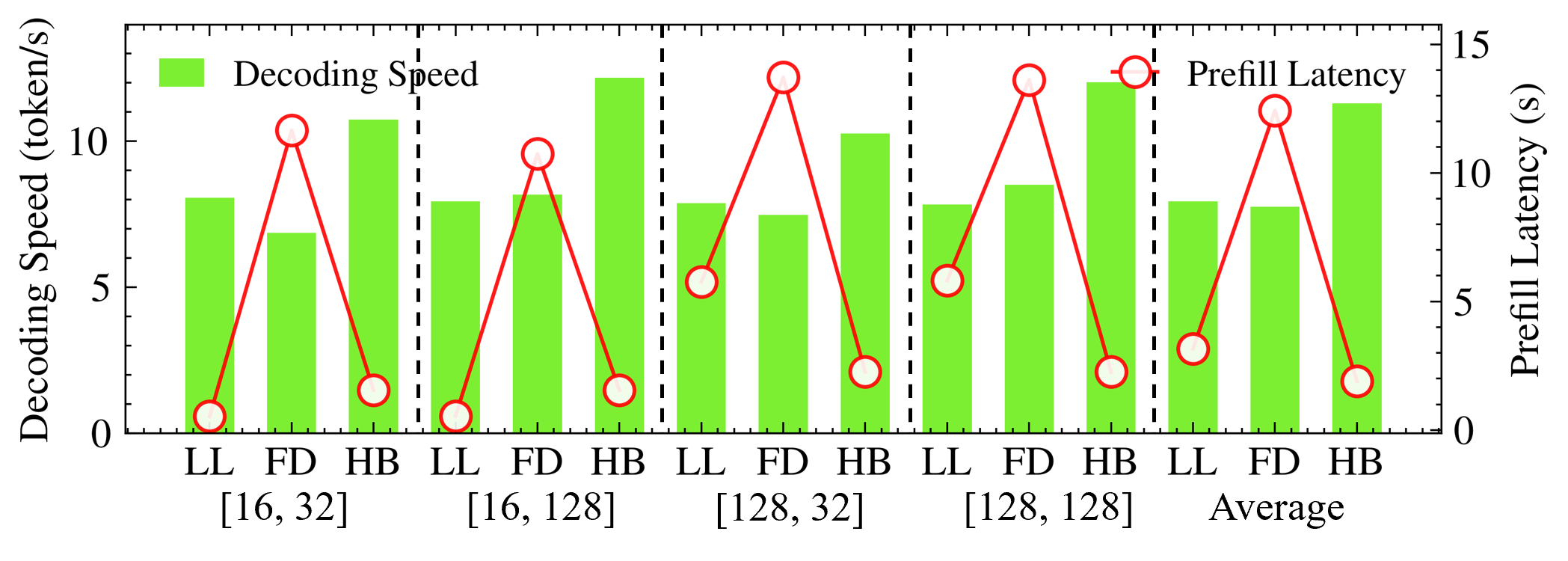}}
    \caption{Inference speed for RTX 4090 + CPU setup.}
  \label{fig:cpu-compute}
\end{figure}

Although we primarily focus on common GPU-centric scenarios, sometimes there are enough CPU resources to help speed up the computation. To accommodate such cases, we implement a CPU-GPU cooperative computing mode to enhance system flexibility and maximize hardware resource utilization. We evaluate this computing mode using the RTX 4090 + CPU setup in Table~\ref{tab:compare-settings}.

From the results in Figure~\ref{fig:cpu-compute}, we can observe that \modelname consistently outperforms Llama.cpp in decoding speed. For prefill latency, \modelname only uses the GPU during the prefill stage, which allows it to perform well with longer prompts (128 tokens) but shows reduced performance with shorter prompts (16 tokens). On average, \modelname achieves lower prefill latency than Llama.cpp. 
Specifically, \modelname delivers a 1.31x and 1.42x speedup in decoding speed, along with a 25\% and 40\% reduction in prefill latency for the two evaluated models, respectively.

Fiddler shows a very slight advantage on Mixtral-8x7B compared to our system in decoding speed, which stems from Fiddler's faster CPU computation speed. Fiddler processes the same expert in 3ms, compared to 5ms for our system, due to differences in implementation (Fiddler uses PyTorch’s interface, while our system relies on Llama.cpp’s interface). However, \modelname surpasses Fiddler on Phi-MoE, where the smaller expert size leads to similar CPU computation speeds for both interfaces. Additionally, Fiddler’s prefill latency increases exponentially with the number of experts, which affects its performance with Phi-MoE, where the number of experts is double that of Mixtral-8x7B, resulting in much higher prefill latency for Fiddler. As a result, Fiddler’s advantage is limited to specific situations and is not well-suited for different models and environments. Compared to Fiddler, \modelname achieves a 0.99x and 1.46x speedup in decoding speed, and a 42\% and 85\% reduction in prefill latency for the two evaluated models, respectively. 

Although utilizing the CPU as a helper is not the primary scenario for \modelname and it doesn’t experience the long expert-loading delays typical of other expert-offloading systems, \modelname still provides better performance than current systems, even if the benefits are smaller compared to GPU-centric computation scenarios.

\subsection{Ablation Study}

To verify the effectiveness of each component in \modelnamenospace, we separately evaluate the dynamic 
 expert loading mechanism, adaptive expert prefetching technique, and multidimensional expert caching policy.

\subsubsection{Dynamic Expert Loading Mechanism Analysis}

\begin{figure}[t]
    \centering
    \includegraphics[width=\linewidth]{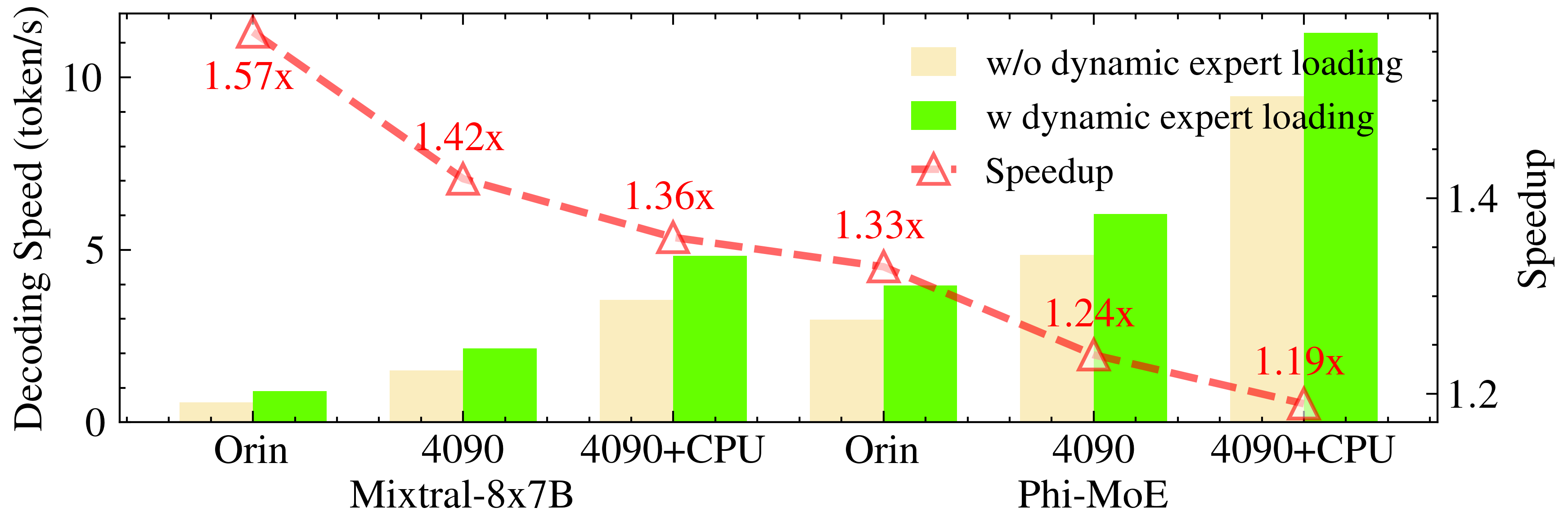}
    \caption{Inference speedup of dynamic expert loading.}
    \label{fig:misshandlerspeed}
\end{figure}

To evaluate the benefits of the dynamic expert loading mechanism, we test models across all setups and averaged the results for different input and output lengths. As shown in Figure~\ref{fig:misshandlerspeed}, using the dynamic loading mechanism provides a speedup ranging from 1.19x to 1.57x under different configurations. The highest speedup is observed when running models on the Jetson Orin device, which is due to its relatively slower data transfer speeds compared to the RTX 4090. Conversely, the lowest speedup is seen on the RTX 4090 assisted by the CPU, as in this setup, the performance gains primarily stem from CPU computing improvements with low-precision experts, rather than faster expert loading. Additionally, we observe that the Mixtral-8x7B model achieves a greater speedup than the Phi-MoE model, likely due to the larger expert sizes in the Mixtral-8x7B model. Overall, these findings suggest that the dynamic expert loading mechanism is especially advantageous in environments with slower data transfer speeds and for models with larger experts.

\subsubsection{Adaptive Expert Prefetching Technique Analysis}

\begin{figure}[t]
    \centering
    \subfloat[Gating module cost]{\includegraphics[width=\linewidth]{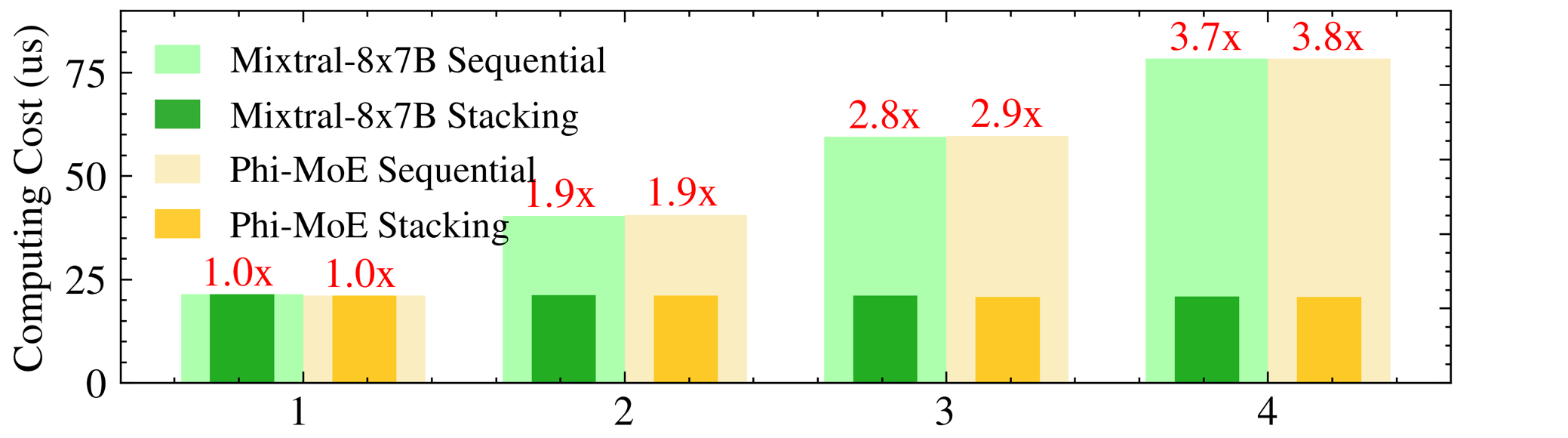}}
    
    \hfill
    \subfloat[Model inference speed]{\includegraphics[width=\linewidth]{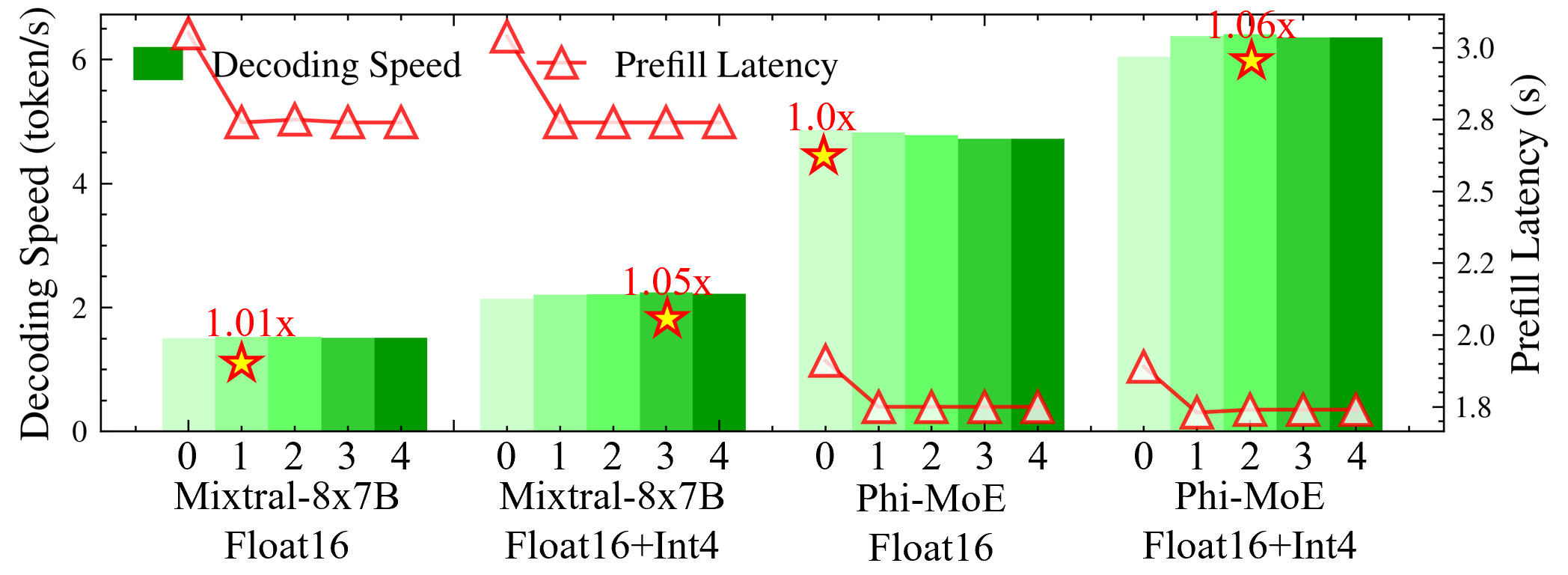}}
    \caption{Gating module cost and model inference speed with different configurations of the adaptive predictor, where 0 means no prediction, and 1, 2, 3, and 4 represent predictions for 1, 2, 3, and 4 subsequent layers, respectively.}
  \label{fig:predictor-analysis}
\end{figure}

We first evaluate the stacking operation in our design. As shown in Figure~\ref{fig:predictor-analysis}-(a), as the number of predicting layers increases, the sequential operation time increases linearly, whereas our stacking operation remains stable, confirming our previous analysis. Next, to assess the benefits of the adaptive expert prefetching technique, we conducted experiments on the RTX 4090 setup, comparing performance with and without the dynamic expert loading. As illustrated in Figure~\ref{fig:predictor-analysis}-(b), the prefetching technique provides notable advantages during the prefill stage, reducing prefill latency by approximately 10\% across all cases. This is because, in most cases, the prefill stage utilizes all experts of each layer, resulting in 100\% prediction accuracy. However, during the decoding stage, the prefetching technique offers only modest  benefits due to the high penalties incurred from incorrect predictions, aligning with previous analysis in Section~\ref{sec:predictor-design}. Without the dynamic expert loading (Float16), the prefetching technique yields minimal  performance improvements (1.01x speedup for Mixtral-8x7B) and even slightly degrades processing speed for Phi-MoE, likely due to slightly lower prediction accuracy.  However, when combined with the dynamic expert loading (Float16+Int4), the prefetching technique begins to show its value, offering a speedup of approximately 1.05x in both Mixtral-8x7B and Phi-MoE models. Additionally, we observe that the value of $p$ influences performance and larger values of $p$ may lead to slightly worse outcomes. Based on our tests, we recommend setting $p$ in range 1 to 3. Overall, with our dynamic precision expert loading method, the adaptive expert prefetching technique can fully capitalize on its potential.

\subsubsection{Multidimensional Expert Caching Policy Analysis}

\begin{figure}[t]
    \centering
    \subfloat[Different cache policies]{\includegraphics[width=0.49\linewidth]{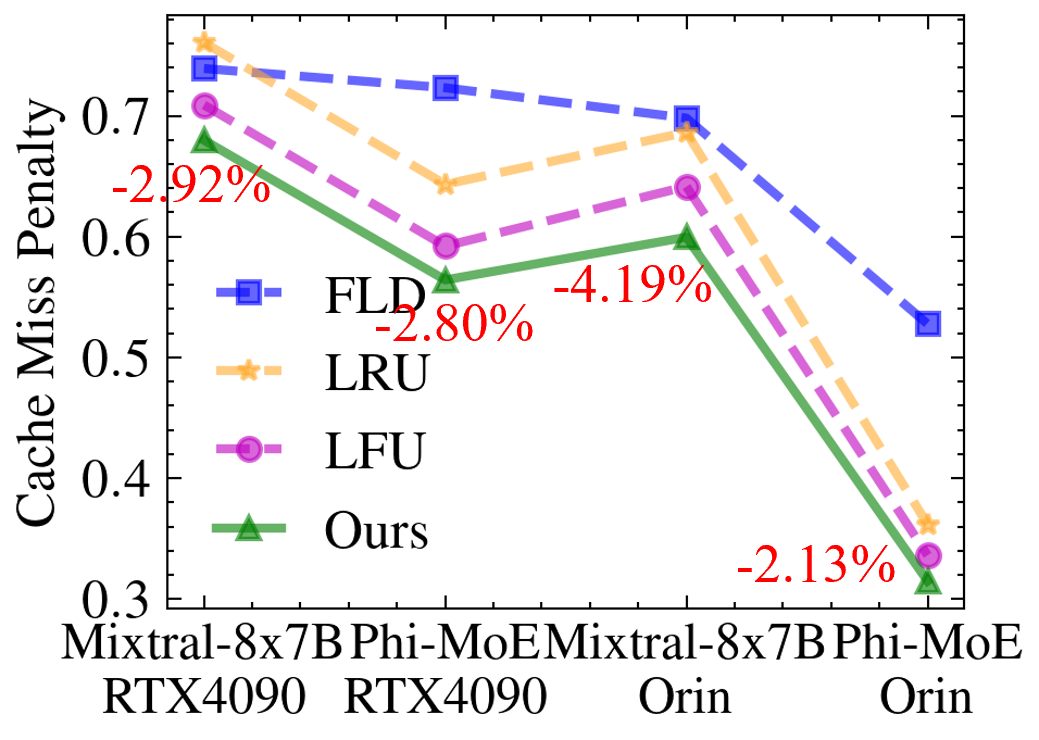}}
    \subfloat[Model level  vs sequence level]{\includegraphics[width=0.49\linewidth]{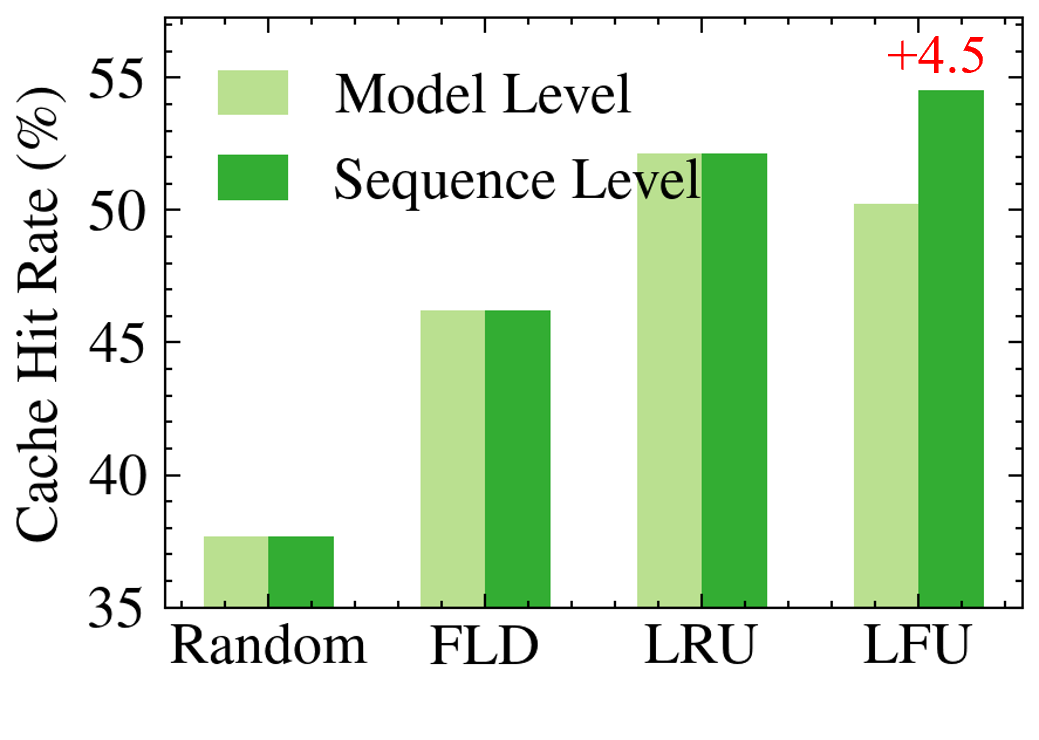}}
    \caption{Cache performance for different strategies.}
  \label{fig:cache-policy-verify}
\end{figure}

To verify the robustness of our proposed cache replacement policy, we compared it with other commonly used policies across different settings. As shown in Figure~\ref{fig:cache-policy-verify}-(a), our policy consistently achieves the lowest cache miss penalty (normalized with random policy baseline, with the reduction compared to LFU highlighted, as LFU is the best among the baselines). Although the improvements may be small in some conditions, our policy is flexible and reliable across various models and environments, while other methods exhibit inconsistent performance. For example, LRU works well for the Phi-MoE model on the Orin device, but performs poorly for the Mixtral-8x7B model on the same device. Overall, our policy shows a $4.69\% \sim 8.68\%$ cache miss penalty reduction over LRU and a $2.13\% \sim 4.19\%$ reduction over LFU. Additionally, we compared the model-level and sequence-level performance of different policies, as shown in Figure~\ref{fig:cache-policy-verify}-(b). The sequence level primarily affects LFU policies,  where sequence-level LFU gains a 4.5\% increase in hit ratio. This behavior is expected, as the balanced loss function in the training process of MoE-based LLMs encourages equal frequency of expert selection, as discussed earlier. In contrast, other policies perform similarly across both model-level and sequence-level conditions. Therefore, a sequence-level policy is more suitable for MoE when considering the LFU strategy.

\section{Related Work}

\noindent \textbf{Expert-offloading systems for MoE-based LLMs.}
Given the significant GPU memory requirements of MoE-based LLMs, many systems have been developed to optimize inference on memory-constrained devices by offloading expert parameters to CPU memory or SSD. Some systems, such as MoE-Offloading~\cite{eliseev2023fast}, MoE-Infinity~\cite{xue2024moe}, Pre-gated MoE~\cite{hwang2024pre}, and SwapMoE~\cite{kong-etal-2024-swapmoe}, focus on optimizing prefetching techniques and cache replacement policies. Others, like EdgeMoE~\cite{yi2023edgemoe} and AdapMoE~\cite{adamoe}, aim to reduce expert-loading costs. Additionally, approaches such as Fiddler~\cite{kamahori2024fiddler} leverage CPU computational power to assist in the inference process. However, these systems do not fully exploit the key characteristics of MoE-based LLMs, resulting in large expert-loading costs or big drops in model accuracy on edge devices.

\noindent \textbf{Optimized inference systems for LLMs.} 
The large size and complexity of LLMs have driven the development of various inference systems aimed at optimizing the efficiency and performance of the inference process. Some of these works focus on adaptive offloading, such as VLLM~\cite{kwon2023efficient}, DeepSpeed~\cite{deepspeedinference}, Accelerate~\cite{accelerate} and Llama.cpp~\cite{llamacpp}, which speed up inference by offloading model weights or internal activations to alternative hardware, such as CPU memory or SSDs. Other works focus on exploiting the sparsity in LLMs to reduce computational cost and memory consumption by skipping the computation of inactive neurons. Examples include Deja Vu~\cite{liu2023deja}, PowerInfer~\cite{song2023powerinfer, xue2024powerinfer}, and CATS~\cite{lee2024cats}. However, these systems have primarily been designed to target dense LLMs, where all model parameters are utilized uniformly across different inputs. As a result, they often fail to exploit the unique characteristics of MoE-based LLMs.

\noindent\textbf{Model compression techniques for LLMs.} 
Model compression is a promising area of research that offers various techniques to effectively deploy LLMs in resource-constrained environments. The core idea of model compression is to reduce the size of a large model while preserving its performance. Popular compression techniques, such as quantization~\cite{wang2023bitnet, frantar2022gptq, dettmers2022gpt3, xiao2023smoothquant}, network pruning~\cite{men2024shortgpt, li2023losparse, ma2023llm, sun2023simple}, knowledge distillation~\cite{gu2023knowledge, agarwal2023gkd, huang2022context}, and low-rank factorization~\cite{li2023merge, wang2024svd, yuan2023asvd}, enable the creation of more resource-efficient LLMs. 
However, these methods only focus on algorithmic optimization, so they often overlook the practical challenges of deploying LLMs on resource-constrained devices, leading to inefficient deployments.

\section{Conclusion}

In this work, we introduce \modelnamenospace, a flexible  and efficient inference system for deploying MoE LLM models on memory-constrained edge devices. By addressing the high cost associated with expert loading in existing MoE inference systems, \modelname enables significant speedups in inference performance. Additionally, we have integrated \modelname into the Llama.cpp inference framework. This integration allows for flexible scalability across a wide range of platforms. Overall, \modelname represents a significant step forward in making complex MoE models more accessible and practical for edge computing environments.

\bibliographystyle{plain}
\bibliography{sample-base}
\end{document}